\PassOptionsToPackage{dvipsnames}{xcolor}
\documentclass[sigconf,nonacm]{acmart}

\definecolor{acGray70}{rgb}{0.65,0.65,0.65}
\definecolor{acGray30}{rgb}{0.85,0.85,0.85}
\definecolor{acGray45}{rgb}{0.775,0.775,0.775}
\definecolor{acGray35}{rgb}{0.825,0.825,0.825}
\definecolor{acBlack80}{rgb}{0.20,0.20,0.20}
\definecolor{acTealDark}{rgb}{0.0,0.40,0.40}

\setcopyright{none}
\renewcommand\footnotetextcopyrightpermission[1]{}

\usepackage{amsmath}
\usepackage{tabularx}
\usepackage{tcolorbox}
\usepackage{listings}
\usepackage{todonotes}
\usepackage{placeins}
\usepackage{appendix}
\usepackage{multirow}

\usepackage{bbm}

\newcommand{\indicator}{\mathbbm{1}}
\newcommand{\tol}{\eta}
\newcommand{\tabnotes}[1]{%
	\par\smallskip\noindent{\footnotesize\textit{Notes.} #1}\par%
}
\newcommand{\dropcap}[1]{#1}

\tcbuselibrary{listings,skins,breakable}

\newcommand{\rolehead}[1]{%
	\noindent{\color{acGray70}\sffamily\small\bfseries #1}\enspace
	{\color{acGray30}\leaders\hrule height 0.4pt\hfill\kern0pt}%
}
\newcommand{\systag}{\rolehead{System}}
\newcommand{\usertag}{\rolehead{User}}

\newtcblisting{promptbox}[1][]{
	enhanced,
	breakable,
	colback=white,
	frame hidden,
	borderline west={2.5pt}{0pt}{acGray45},
	left=10pt, right=4pt, top=6pt, bottom=6pt,
	title={\small\sffamily\bfseries #1},
	colbacktitle=white,
	coltitle=acBlack80,
	titlerule=0.4pt,
	titlerule style={acGray35},
	listing only,
	listing options={
		basicstyle=\ttfamily\footnotesize,
		breaklines=true,
		breakindent=0pt,
		breakatwhitespace=true,
		columns=fullflexible,
		keepspaces=true,
		showstringspaces=false,
		tabsize=4,
		escapeinside={(*}{*)},
		emph={COMPETITION_DEPTH,N_PARTICIPANTS,MIN_RANGE,MAX_RANGE,OPPONENT_TYPE,PAYOFF_TABLE},
		emphstyle=\color{acTealDark},
		literate=
		{\{}{{\textcolor{acTealDark}{\{}}}1
		{\}}{{\textcolor{acTealDark}{\}}}}1,
	}
}

\begin{document}

\title{The Fragility of Strategic Thinking in Large Language Models}

\author{Enric Junqu\'e de Fortuny}
\email{enric@skku.edu}
\affiliation{%
  \institution{SKK Graduate School of Business, \\Sungkyunkwan University}
  \city{Seoul}
  \country{South Korea}
}

\author{Veronica Roberta Cappelli}
\email{vcappelli@iese.edu}
\affiliation{%
  \institution{Department of Managerial Decision Sciences, \\IESE Business School}
  \city{Barcelona}
  \country{Spain}
}


\begin{abstract}
\medskip
\noindent
Large Language Models (LLMs) are increasingly applied to domains that require reasoning about other agents' behavior, such as negotiation, policy design, and market simulation. However, can we trust LLMs to think strategically in complex situations? Existing research mostly evaluates LLMs' adherence to equilibrium play or their exhibited depth of reasoning, leaving open whether they display \textit{strategic thinking} meant as the ability to form coherent conjectures about other agents, to evaluate possible actions conditional on those conjectures, and to best respond to them. We develop a framework to identify this ability by disentangling belief formation, evaluation, and choice in static complete-information games across a series of non-cooperative environments. By jointly analyzing models' revealed choices and reasoning traces, and introducing a new context-free game to rule out imitation from memorization, we show that strategic thinking in current frontier LLMs is real but fragile: models execute best responses to exogenous conjectures and form opponent-contingent conjectures when left unconstrained. Yet under increasing complexity explicit recursion gives way to model-specific logic shifts and heuristic rules of choice, both within and outside equilibrium reasoning. Further, these heuristics do not map directly onto the systematic biases typically observed in human strategic behavior. These findings, already emerging in noiseless settings, warrant caution in the application of LLMs as strategic agents in complex environments.\\
\\
\noindent\textbf{Significance Statement.}\enspace
Artificial intelligence systems are increasingly deployed in high-stakes agentic settings. From autonomous negotiations and military decision support to autonomous financial trading, success hinges on anticipating what others will do.
Yet it remains unclear whether these systems genuinely reason about opponents or merely pattern-match. We develop a method to test whether large language models can form conjectures about other agents' behavior, evaluate their implications, and choose optimal responses.
We find that frontier models exhibit responses coherent with strategic thinking, but, under greater complexity, explicit recursion gives way to model-specific logic shifts that depart from strategic thinking.
Our results highlight the need to better understand how language models assess and anticipate the behavior of other agents in high-stakes interactive settings.
\end{abstract}

\maketitle
%\linenumbers

	\dropcap{I}n February 2025, a Washington D.C. think-tank released a ``Ukraine-Russia Peace Agreement Simulator'' to Western government negotiators~\cite{_AIModelsCould_2025}. This Large Language Model (LLM) was designed to draft treaties based on users' outcome preferences and likelihood of adoption. This is not the only example of applying LLMs to high-stakes multi-agent settings. LLMs are used in decision-making scenarios in a wide variety of processes, including financial trading algorithms, automated contract negotiation and analysis, and policy-making simulations~\cite{xiao_TradingAgentsMultiAgentsLLM_2025,narendra_EnhancingContractNegotiations_2024,coz_WhatWouldLLM_2025}. However, a meaningful implementation of these applications requires agentic systems to exhibit not only generic reasoning abilities to identify alternatives, but also, and more importantly, the ability to plan and choose a course of action, while getting in the shoes of other decision makers involved, that is, to think strategically.
	
	Can LLMs actually engage in strategic thinking? LLMs are trained on next-token prediction over massive language corpora, and smaller models remain limited to this objective. Remarkably, new capabilities emerge in larger models that are absent in smaller ones~\cite{wei_EmergentAbilitiesLarge_2022}. One of the most interesting phenomena observed in LLMs is the emergence of problem-solving abilities across domains such as math and economic decision-making. For example, {\em Chain-of-Thought prompting} revealed that LLMs could be steered toward step-by-step reasoning behaviors, an ability not exhibited by earlier models~\cite{wei_ChainofThoughtPromptingElicits_2022}. Encouraged by these early findings, researchers began aligning models toward reasoning at both training and inference stages (e.g., GPT-4 o1, DeepSeek v3; \cite{lee_EmergenceStrategicReasoning_2025}). Today's frontier models perform on par with human experts on several structured tasks, as demonstrated by advanced problem-solving benchmarks and the recent successes in the Math Olympiads. Although problem solving is quickly becoming one of the key defining features of LLMs, it is still unclear whether this behavior implies or is equivalent to strategic thinking.

	Classical game theory identifies three critical components of strategic thinking: forming beliefs about others, evaluating options, and choosing best responses~\cite{battigalli_GAMETHEORYAnalysis_2023}. Strategic thinking captures the idea of \textit{planning} and \textit{choosing} the best feasible course of action given the available information. Crucially, it also involves knowing that each decision-maker's payoff depends on the choices of all other players involved. Strategic thinking therefore requires additional layers beyond the observed economically rational choice behavior exhibited in GPT~\cite{chen_EmergenceEconomicRationality_2023}. 
	To exhibit strategic thinking, LLMs must not only plan a sequence of (economically rational) actions in a given task environment, but also effectively use information about other agents.

	In the past few years, much research has explored strategic thinking in LLMs, typically by looking at consistency with equilibrium, alignment with human behavior in experimental settings, or proxies for greater reasoning depth, rather than examining LLMs' thought process in conjunction with their behavior.
	
	A growing body of research has examined LLMs in strategic settings, producing important insights into their capabilities and limitations. One line of work measures how closely models approximate game-theoretic benchmarks such as Nash equilibrium or level-$k$ reasoning, revealing that frontier models can achieve considerable strategic sophistication \cite{chen_EmergenceEconomicRationality_2023, horton_LargeLanguageModels_2023,duan_GTBenchUncoveringStrategic_2024}, though the picture is nuanced. In particular, context sensitivity is contested, with some studies finding strong framing effects \cite{lore_StrategicBehaviorLarge_2024, horton_LargeLanguageModels_2023} and others reporting rigid, model-specific heuristic application \cite{zheng_NashEquilibriumBounded_2025}; cognitive decision making biases from training data are partially replicated but often diverge from human patterns in structure \cite{suri_LargeLanguageModels_2023,echterhoff_CognitiveBiasDecisionMaking_2024,macmillanscott2024irrationality,bini2026behavioral}. 
	However, by definition, high depth of reasoning or equilibrium play {\em per se} does not imply strategic reasoning. In fact, it can be trivially shown that, for a player, choosing an action that is compatible with an arbitrarily great depth of reasoning in a strategic setting is not unconditionally optimal \cite{battigalli_GAMETHEORYAnalysis_2023}. Analogously, if an LLM "unilaterally" plays a Nash equilibrium strategy, it may not be a best response to the (belief that a player would make a) low depth reasoning choice. A related distinction appears in algorithmic game theory, where learning to play well against a bounded opponent requires modeling that opponent's behavior, not merely applying a fixed solution concept for the stage game~\cite{freund_EfficientAlgorithmsLearning_1995}.

	A complementary line of work focused on studying strategic interactions in canonical games, showing mixed results where unconstrained LLMs have been found to be compatible with both shallower and greater depths than humans, yet converge toward equilibrium under repetition \cite{lu2025beauty,alekseenko2025strategizing,yang_LLMsKnowWhen_2025} and have been found to be steerable through prompt design \cite{zhang_KLevelReasoningEstablishing_2024, kempinski2025game, manning_GeneralSocialAgents_2025}.
	These contributions have substantially advanced our understanding of \textit{what} LLMs do in strategic environments. A natural but largely unexplored next step is to investigate \textit{why}. This entails moving beyond outcome-level classification and examine the antecedents of strategic choice: whether models form coherent conjectures about their opponents, evaluate the implications of those conjectures, and select actions accordingly and correctly.
	To this end, we develop a method to disentangle the three components of strategic thinking (beliefs, evaluation and choice) and implement it in a series of experiments to study whether different classes of LLMs exhibit \textit{strategic thinking} behavior.
	As such, this paper develops a framework for identifying a specific microfoundation of strategic behavior in LLMs under controlled conditions. Existing evaluations often classify LLM behavior using equilibrium or bounded-rationality benchmarks. We instead identify whether choices are contingent on coherent conjectures about others, and show that frontier models can both act on supplied conjectures and form differentiated opponent-specific conjectures.
	We use a series of non-cooperative games with complete information that vary in their degree of both abstraction and complexity of choice: the Beauty Contest Game (henceforth, BCG), the 11–20 Money Request Game (henceforth, MRG), and a novel Unlabeled Matrix Game (henceforth, UMG). These settings allow us to infer antecedents of choice behavior unambiguously, since players do not need to hold proper \textit{beliefs} about basic elements of the game, as they are endowed with complete information about it\footnote{A disambiguation note: in what follows, for the reason just specified and with a slight abuse of terminology, we will be looking at the impact of players {\em conjectures}, which are a (probabilistic) belief about the behavior of other players. Whereas the term {\em belief} is used in the literature to refer to a more general type of uncertainty about elements of the game \cite{battigalli_GAMETHEORYAnalysis_2023}. %The two notions are formally equivalent in our classes of games.
	} and there are no confounds stemming from multi-period payoff analysis. This design allows us to explore whether the exhibited modes of reasoning transfer across contexts. We measure consistency of LLMs' choices with theoretical predictions, systematically varying agents' \textit{conjectures} on opponents' play. We complete our study with the models' declared Chain-of-Thought to identify behavioral consistency, points of failure, and categorize reasoning processes. 

	We find that LLMs regularly exhibit best-response behavior across three types of strategic interactive situations and are able to do so at arbitrarily targeted levels of depth of reasoning. When provided with an exogenous conjecture about their opponents' reasoning, models derive its implications and select choices that are coherent with economic rationality. Moreover, LLMs independently formulate such conjectures about the behavior of human and synthetic agents by displaying coherent meta-thinking, conditioning their choices on the identity of the opponent, and, as a result, often self-constraining their depth of reasoning. %\vrc{add overthinking? also caveat} 
	In more complex settings, characterized by a lack of convergence of best response behavior, we observe systematic shifts in reasoning logic where recursive belief modeling is at times substituted with equilibrium play or heuristic shortcuts in decision making, intended as simplified decision rules. 
	Our findings provide direct evidence that strategic thinking, as distinct from reasoning depth or memorization, can emerge from language model-based systems, thus contributing to our understanding of the representational value of language as vehicle for operationalizing formal concepts.

	Our analysis of action choice, jointly with Chain-of-Thought, identifies the antecedents of these deviations from iterative best-response behavior: model-specific logic shifts that produce emergent heuristics, either on their own or layered onto equilibrium-compatible play. This adds a novel dimension to research on whether LLMs replicate aspects of human cognition~\cite{suri_LargeLanguageModels_2023,echterhoff_CognitiveBiasDecisionMaking_2024,saeedi_HeuristicsBiasesAI_2025,lee_EmergenceStrategicReasoning_2025,zheng_NashEquilibriumBounded_2025}.

	More fundamentally, the observed behavior of LLMs in relatively more complex strategic situations warrants caution in the adoption of LLMs as strategic agents. In fact, these logic shifts are not implications of strategic reasoning. This observation raises the question as to whether they represent attempts at providing a deterministic and unique action choice among those available when best responses are not well defined and whether the mechanism underlying these deviations is stable, and thus its outcome predictable across contexts, at least within models \cite{kalai2025language}. 
	Our study therefore highlights the need for further research on emergent strategic thinking in LLMs across structured and unstructured environments to ensure their suitability for strategic applications. The urgency of this question is underscored by the fact that LLM-based agents are already being deployed autonomously in commercial negotiations and financial trading at scale~\cite{pactum2024,hsgac2024}.

	\section{Methods}
	
	\subsection*{Conceptual framework: beliefs, best responses, and equilibria}

	In this section, we provide the mathematical definition of the main game theoretic concepts we use throughout the paper. In particular, we use the term \textit{game} to denote the mathematical structure that fully describes the interaction among players in a competitive setting. Since our focus is the analysis of strategic thinking of LLMs, we highlight the notion of best response, identified as the ability of a player to choose the most advantageous course of action, given their probabilistic assessment of the actions that their opponents may take. This encapsulates the simultaneous ability to hold a conjecture about opponents' behavior and act strategically upon it. Relatedly, we recall the definitions of widely used behavioral benchmarks to classify observed behavior: depth of reasoning and Nash equilibrium. Qualitatively, depth of reasoning relates to the ability of a player to iteratively put themselves into the other players' shoes, thus forecasting their action choice. Whereas a Nash equilibrium characterizes a profile of actions such that no player has an incentive to act differently, given the other players' choice, and thus represents a stable solution. However, as remarked in the introduction, neither Nash equilibrium play nor arbitrarily great depth of reasoning is a standalone marker of strategic thinking, as the latter necessarily requires an evaluation of the prospective opponents' behavior. 
	
	\paragraph{Analysis of Strategic Thinking}
	Each interactive decision problem is described by the following elements: (i) a set of players $I$; (ii) for each player $i\in I$ a set $A_i$ of actions that can be chosen by $i$ in the game; (iii) a set of outcomes $Y$; (iv) an extensive form $\mathcal{E}$, representing the rules saying whose turn it is to move, what a player knows, and what are the feasible actions at each point of the game, determining a set $Z$ of sequences of feasible actions; (v) an outcome function $g:Z\rightarrow Y$ specifying an outcome for each sequence of feasible actions; (vi) $(v_i)_{i\in I}$, where each $v_i:Y\rightarrow \mathbb{R}$ is a utility function representing player $i$'s preferences over outcomes. 
    We call \textit{game} the tuple $G=\langle I,(A_i)_{i\in I},Y,\mathcal{E},g,(v_i)_{i\in I}\rangle$ describing an interactive decision problem. 
     We will focus on static games of complete information, that is, interactive decision problems where it is common knowledge 
	that $G$ is the game to be played. Given these restrictions, we will use the terms action and strategy interchangeably and $Z = A = \bigtimes_{i \in I} A_i$. Hence, from (v) and (vi), we can obtain, for each player $i \in I$, a payoff function $u_i := v_i \circ g: A \rightarrow \mathbb{R} $ that assigns to each action profile $a \in A$ player $i$'s utility $v_i(g(a))$ from the consequence $g(a)$.
    
	Note that choices in game theory can, in principle, be regarded as stochastic. Let us consider the case of players being drawn from a large population in which a certain fraction chooses an action among the available ones. The population's action $\alpha$ is a mixture of the actions chosen by players in the population, weighted by the fraction of players that chose it. This mixture, formalized as a probability distribution over actions, is, in game theory, a {\em mixed action}.\footnote{According to a different interpretation, it can be the case that a player, instead of choosing an action, could delegate his decision to a probabilistic device, assigning a probability to each available action~\cite{battigalli_GAMETHEORYAnalysis_2023}.}

	\paragraph{\em Best responses and conjectures.}
	Central to the notion of strategic thinking is the notion of best response. However, in static games, players cannot observe the actions of the other players before making their own choice due to the fact that moves are simultaneous. Hence, at the moment of action choice, they need to form a conjecture over their opponents' actions, that is, a (probabilistic) belief about the behavior of other players.
	
	It follows that actions are a best response not to observed behavior but to the conjecture that the player holds about the behavior of other players. An action is a best response if it maximizes the player's expected utility, given their conjecture $\mu^i$, that is 
	\[
	a_i^\text{BR} \in \arg\max_{a_i} u(a_i,\mu^i)
	\]

	\textit{Operational measure.} We evaluate whether the model’s choice $\hat a_i$ is a best response to $\mu^i$. In practice, we compute best-response regret (BRR) as the payoff gap between $\hat a_i$ and the optimal best response. Best-response accuracy is recorded as $\indicator[\mathrm{BRR}_i \le \varepsilon]$, where $\varepsilon=0$ for exact compliance (MRG, UMG) and $\varepsilon=5\%$ for BCG.\footnote{In the canonical BCG with $n$ participants and parameter $p$ (target $= p \times$ mean of all $n$ chosen numbers, including the focal player's), the level-$k$ best response is the fixed-point sequence $x^{(k)} = L_0 \cdot \left[\frac{p(n-1)}{n-p}\right]^{k}$ with $L_0$ the midpoint of the action range \cite{nagel_UnravelingGuessingGames_1995}. We count a choice $\hat x$ as correct if $\lvert \hat x - x^{(k)} \rvert / x^{(k)} \le \tol$ with $\tol=5\%$ unless otherwise noted.}

	\paragraph{\em Depth of reasoning.}
	Conjectures about other players' behavior are based on a variety of factors and, in theoretical analysis, are treated as exogenous. A large body of empirical and theoretical work looked at the implication of holding specific conjectures to provide structure to deviations from equilibrium play. In particular, level k and cognitive hierarchy theory formulate hypotheses around the extent of strategic sophistication\footnote{By {\em strategic sophistication} here we refer to the degree to which a player incorporates the elements of a game's formal structure and the incentives of other players when deciding on their strategy.} 
	of players, thus pinning down players' conjectures. According to level k, a level 0 player is defined as non-strategic and is assumed to choose a strategy $\bar{a}$ at random among available ones. A level 1 player conjectures that his opponents will choose strategy $\bar{a}$ and thus selects the strategy $a_j^1$ that best responds to it. Reasoning iteratively, a level k player conjectures that his opponents will choose strategy $a_{-i}^{(k-1)}$ and thus selects the strategy $a_i^k$ such that 
	\[
	a_i^k \in \arg\max_{a_i} u(a_i,a_{-i}^{k-1})
	\]
	
	Similarly, the cognitive hierarchy theory, instead assumes that players reason at all levels up to \( k-1 \), where k is distributed according to a Poisson with mean \( \tau > 0\), that is
	\[
	f_k(j;\tau) = \frac{e^{-\tau} \tau^j}{j!} \Bigg/ \sum_{m=0}^{k-1} \frac{e^{-\tau} \tau^m}{m!}
	\].
	
	\textit{Operational measure.} When a prior is imposed by us (targeted-depth variants), we check whether the model's final choice coincides with $a_i^{(k)}$ (within the BCG tolerance when numeric). In open-ended variants, we estimate the implied depth (either $k$ or $\tau$) from the distribution of observed choices and traces.
	
	\paragraph{\em Nash equilibrium.}
	
	A Nash equilibrium is an action profile such that each player is best responding to the conjecture they hold about other players' behavior, and this conjecture is correct. We denote by $S_i \subseteq A_i$ the set of equilibrium actions for player $i$ in the case of pure equilibria. In the case of mixed equilibria, we denote by $S_i^+ \subseteq A_i$ the support of the equilibrium distribution (and by $S_i^-$ its complement). It is worth noting that the Nash equilibrium is a fixed point of the k-level reasoning outlined above. It is such that player i assumes that other players will respond best to $a^*_i$ and thus play $ a^*_{-i} \in \arg\max_{a_{-i}\in A_{-i} } u(a^*_i,a_{-i})$ and therefore player i plays $ a^*_{i} \in \arg\max_{a\in A_{i} } u(a,a^*_{-i})$. Importantly, despite this fact, k-level reasoning does not always converge to a Nash equilibrium. In all the strategic decision situations analyzed in the present paper, the existence of Nash equilibria is guaranteed, although not always in pure strategies.
	
	\textit{Operational measure.} 
	For pure equilibria (BCG, UMG), we record whether $\hat a_i \in S_i$, where $S_i \subseteq A_i$ is the set of equilibrium actions for player $i$. For mixed equilibria (MRG), we record \emph{support coverage} $\indicator[\hat a_i \in S_i^+]$, where $S_i^+$ denotes the support of the equilibrium distribution (and $S_i^-$ its complement). Across repeated trials, we also compare the empirical action distribution $\hat P_i$ to the theoretical equilibrium distribution by reporting entropy and distance metrics.

	\subsection*{Models and settings}

	Table~\ref{tab:models} (Appendix) shows the different models under investigation, grouped by type. We include frontier reasoning models as identified by high performance on GPQA-D and HLE, standard instruction-following models, and a diffusion-based language model (Mercury~2), since the latter two categories are privileged in agentic applications that require fast and efficient computation. 
	All model settings are set to their respective defaults except for the temperature, which we set to $t=0.75$, in order to balance consistency and heterogeneity of the resulting behaviors. Next, we describe the three games and how we request the LLMs to make interactive decisions by describing the structure of prompts, or \textit{task environments}, here intended as a particular parametrization of a game and instruction set, in response to which the LLM outputs its choice and its declared reasoning.

	\subsection*{Games}

	\paragraph{Game 1: The Beauty Contest Game (BCG)}
	We begin our analysis with a classic textbook game originally known as {\em the Keynesian beauty contest}~\cite{nagel_UnravelingGuessingGames_1995}. In this game, $n$ players have to simultaneously select a number in the closed interval $[0,100]$. The player whose selected number is closest to a given fraction of the mean of all selected numbers wins a positive prize. In other words, the \textit{target} is equal to the mean of the numbers chosen by all players, multiplied by a predetermined parameter $p\in [0,1)$, where $p$ is common knowledge.
	If there is a tie, the prize is equally divided between the winners. All players whose answers are farther away receive nothing. 
	To gain further insights about the impact of information about opponents on beliefs' formation and strategic choice behavior in LLMs, we introduce variations in the identity of the opponent in the game through the text of the prompt. This game has been selected to provide continuity with respect to current research on LLMs gameplay and due to the simple properties of its equilibria and best response structure.

	\paragraph{Game 2: The 11-20 Money Request Game (MRG)}
	
	In the original version of the 11-20 Money Request Game~\cite{arad_1120MoneyRequest_2012}, two players must each request an amount of money. The amount must be an integer between 11 and 20 shekels. Each player will then receive the amount they request. In addition, a player will receive an additional amount of $B=20$ shekels if they ask exactly for one shekel less than the other player. 

	In two separate experiments, we further investigate the impact of assigning an identity to the LLMs' opponent on strategic choice behavior. This game, also recurrently used in the literature, has been selected to study dynamics of choice behavior in absence of an obvious way of playing,

	since its iterated best-response dynamics do not converge in pure strategies. In particular, because no pure-strategy Nash equilibrium exists,  although a unique mixed-strategy equilibrium does.

	\paragraph{Game 3: The Unlabeled Matrix Game (UMG)}

	While the previous two games are now classics in the game theory literature, we introduce a new generic finite simultaneous-moves game with no background story (UMG). In the UMG, two players ($i$ and $j$) each have to choose one among a finite set of available actions, $A_i$ and $A_j$, respectively. The matrix of payoffs resulting from each possible choice of action profile $(a_i,a_j)$ where $a_i \in A_i = \{A \ldots F\}$ and $a_j \in A_j = \{K \ldots P \}$, is predetermined and common knowledge. In the present paper we use the payoff matrix in Table~\ref{tab:UMG} defining the UMG. Each LLM receives the payoff matrices and is prompted to explicitly walk through each step of recursive reasoning up to their maximum level. Reasoning halts when the LLM has decided that it has converged (as identified by the explicit default "stop"-token in the LLM output). We introduce this game to avoid any possible confound that may arise from the fact that both other games involved are widely known and their properties have been studied in detail.
	
	The three games span the spectrum of prior familiarity: a memorization control we report in Appendix~\ref{sec:appendix-knowledge} shows that BCG is universally recognized by every model in the panel (including the correct lower-bound Nash equilibrium and the Nagel reference), MRG is recognized by name in only a subset of models and with no correct recall of the mixed-strategy support $\{15,\ldots,20\}$, and UMG is universally unrecognized. The convergence of our main findings across all three games therefore separates strategic behavior driven by in-context reasoning from behavior driven by retrieved knowledge.

	\begin{table}[h]
		\centering
		\caption{The Unlabeled Matrix Game payoff matrix.}
		\begin{tabular}{c|cccccc}
			& K  & L  & M  & N  & O  & P  \\
			\hline
			A & (75,75) & (27,27) & (\textbf{\textit{96}},\textbf{\textit{96}}) & (39,39) & (8,8)  & (18,18) \\
			B & ({77},{77}) & (56,56) & (22,22) & (18,18) & ({84},{84}) & (30,30) \\
			C & (72,72) & ({63},{63}) & (41,41) & ({81},{81}) & (48,48) & (77,77) \\
			D & (73,73) & (37,37) & (26,26) & ({82},{82}) & (24,24) & ({92},{92}) \\
			E & (45,45) & (26,26) & ({91},{91}) & (19,19) & ({85},{85}) & (32,32) \\
			F & (58,58) & (48,48) & (83,83) & (67,67) & (25,25) & (\textbf{\textit{94}},\textbf{\textit{94}}) \\
		\end{tabular}
		\tabnotes{Bold-italics = Nash Equilibria (in pure strategies).}
		\label{tab:UMG}
	\end{table}

	\subsection*{Experimental Protocol}
	\paragraph{Task environment: Instructing the model to "choose strategically".} For each game, we instruct the LLMs with the following details: the extensive form, or rules, of the game; the set of players; for each player, the set of actions that can be chosen by them in the game; the set of outcomes contingent on each combination of players' choices of actions; and the players' objectives, that is, to win the game. The LLM is then tasked to select an action among those available, according to the objective provided. For each task environment, we perform 100 runs. At each run, models' choices are recorded in structured format for analysis. All prompts can be found in the appendix. Finally, for all games we have a variant with (e.g., BCG$^T$) and without instruction reasoning traces (e.g., BCG).
	Throughout, our primary claims are identified from revealed choices; reasoning traces are used as secondary diagnostic evidence to interpret the candidate reasoning logics behind those choices. This division mitigates concerns on the possibly limited faithfulness of chain-of-thought to internal computation~\cite{chen_ReasoningModelsDont_2025}.
	
	As a preliminary step, we request the models to explicitly take into account different levels of strategic thinking abilities of their opponents. We measure the extent to which different models are able to handle information about different depths of strategic reasoning. In a subsequent experiment, we request the models to play against humans and explicitly show their reasoning steps (similar to Chain-of-Thought). After that, we analyze their reasoning processes and measure their strategic thinking abilities as measured by best response behavior contingent on their inferred {\em beliefs}.
	
	\section{Results}

	\subsection*{LLMs exhibit best response behavior at arbitrary levels of reasoning depth}
	
\begin{figure}[h]
		\centering
		\caption{Best response accuracy of the different models in the 3 different games (top left, right; bottom left). }
		
		\includegraphics[width=\linewidth]{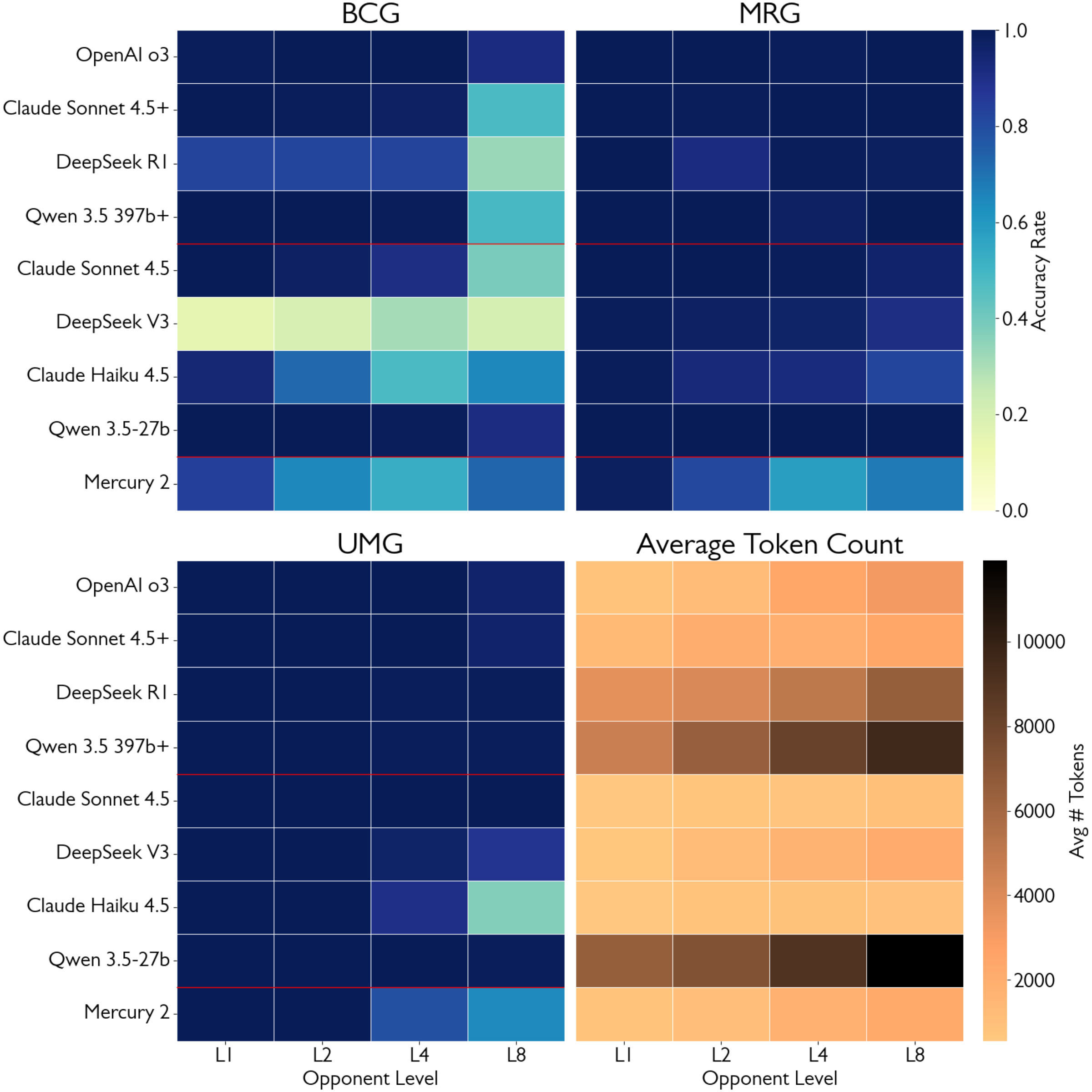}
		\label{fig:model-capability} 
		
		\tabnotes{Bottom right: the number of tokens (including reasoning) for the response, averaged out over all games.}    
	\end{figure}

	A base condition for an LLM to be considered capable of engaging in strategic thinking is its ability to follow through on a given conjecture. That is, once we provide perfect information about the game and exogenously specify the opponent's level of reasoning, we ask whether the model can trace this prior to its logical implication - and up to what depth of thinking. To this end, we instruct the LLM to best respond to increasingly complex, a priori given conjectures about its opponent and observe a clear separation between reasoning and non-reasoning models in their ability to implement best-response behavior~(Figure~\ref{fig:model-capability}).
	
	Most, but not all, models are able to best respond at relatively deep levels of reasoning. We observe that the heterogeneity of performance of models between games is not determined by limits to computational capacity (Fig~\ref{fig:model-capability}, bottom right). Instead, such differences in performance are largely driven by computational accuracy, particularly in the BCG where a closed-form solution exists at every level of depth of reasoning (Figure~\ref{fig:calc-error}, appendix). The BCG-recognition result of the prior-knowledge check (Appendix~\ref{sec:appendix-knowledge}) shows that the Nash limit point is widely recalled, but the level-1/2 conditions that dominate empirical play require in-context arithmetic rather than solution recall, making computational accuracy the binding constraint. This is not a barrier to success in either the MRG or UMG as they do not require floating-point arithmetic to solve. For BCG, increasing relative tolerance thresholds $\tol$ to $10\%$ is often sufficient to meaningfully improve performance relative to the task of best responding to exogenous conjectures about opponents' depth of reasoning, at arbitrarily targeted levels of depth (Figure~\ref{fig:calc-error}, Appendix). This reinforces the idea that in common agentic applications, external calculators are a useful and often necessary tool.

	 %VRC be more explicit about how this is analyzed
	
	\begin{table}[h]
		\centering
		\caption{BCG$^T$ backtracking behavior}
		\begin{tabular}{lccc}
			\toprule
			Model & Mode Level & Total Steps & Overthinking \\
			\midrule
			OpenAI o3             & $L_3$ & $4.0 \pm 0.2$ & $24.0\%$ \\
			Claude Sonnet 4.5+    & $L_3$ & $3.8 \pm 0.1$ & $19.0\%$ \\
			DeepSeek R1           & $L_3$ & $3.5 \pm 0.1$ & $7.4\%$  \\
			Qwen 3.5 397b+        & $L_2$ & $3.1 \pm 0.0$ & $60.6\%$ \\
			\midrule[\heavyrulewidth]
			Claude Sonnet 4.5     & $L_3$ & $4.4 \pm 0.1$ & $16.0\%$ \\
			DeepSeek V3           & $L_4$ & $4.3 \pm 0.1$ & $0.0\%$  \\
			Claude Haiku 4.5      & $L_\infty$ & $4.6 \pm 0.1$ & $36.7\%$ \\
			Qwen 3.5-27b          & $L_2$ & $3.5 \pm 0.1$ & $42.4\%$ \\
			\midrule[\heavyrulewidth]
			Mercury 2             & $L_9$ & $8.0 \pm 0.4$ & $2.0\%$  \\
			\bottomrule
		\end{tabular}
		\label{tab:backtrack}
		\tabnotes{
			Reported values are sample means $\pm$ standard errors (across $n=100$ trials). \emph{Overthinking} is the proportion of trials where the model revisited earlier levels after reaching its final decision.\footnotemark
		}
	\end{table}
    \footnotetext[\value{footnote}]{Throughout, a ``+'' suffix denotes the model's extended-reasoning variant.}

	\subsection*{LLMs self-constrain their depth of reasoning} %\enric{Note, the following section changed substantially:}
	After having observed that many of the models under consideration have the capacity for strategic reasoning in the contexts under analysis, we now turn to measuring the base level of reasoning exhibited in the models. To this end, we remove the exogenously specified opponent and instead let the LLM freely decide how to play. Table~\ref{tab:backtrack} reveals that LLMs frequently {\em overshoot} and then {\em backtrack} on their targeted reasoning depth. This is evidenced both by the statistics (steps taken exceed those required to reach the declared terminal level (Table~\ref{tab:backtrack-full}, Total Steps vs.\ First Terminal), and by the traces themselves (Table~\ref{tab:heuristics_crossgame}, Appendix).

	Most models converge to terminal levels $L_{2\text{--}4}$ despite their demonstrated capacity to reason much further. This self-constraining behavior suggests that LLMs do not treat the game as a purely logical object but instead embed priors about human behavior learned from training data. These priors may reflect statistical regularities in how humans behave in strategic settings and can override normative prescriptions such as Nash equilibrium when the model judges them more contextually relevant. Such behavior reveals an alternative epistemic stance: rather than assuming fully rational opponents, the LLM best-responds to an inferred model of opponent-specific, boundedly rational behavior.
	
	Interestingly, overthinking behavior is consistent within model families suggesting they are an artefact of the training regimes. DeepSeek models show near-zero overthinking (R1: 7.4\%, V3: 0\%), consistent with their reinforcement learning objective of incentivizing efficient reasoning~\cite{deepseekr1_2025}---which may directly discourage the level-revision and hedging more common in models trained with preference optimization such as Claude (Sonnet 4.5+: 19\%, Sonnet 4.5: 16\%, Haiku 4.5: 37\%) and Qwen (high overthinking: 42--61\%). Mercury~2, as a diffusion model, consistently reasons to $L_9$ and shows little to no overthinking (2.0\%), potentially reflecting an architectural difference: diffusion models are not autoregressive and may retroactively revise earlier reasoning to be consistent with a later-committed conclusion.

	\subsection*{LLMs hold conjectures about opponent's type-specific depth of reasoning}
	Evidence of self-constrained depth of reasoning in LLMs suggests
	they can hold different conjectures about different players’ types. We explore this hypothesis in a series of separate experiments where we vary the identity of the opponent. Here, we find that LLMs ascribe different levels of cognitive depth to different player types~(Table~\ref{tab:BCG-bias}). Pooled across the panel, the expert framing shifts the Poisson cognitive-hierarchy parameter upward by $\Delta\hat{\tau}=6.13$ relative to the human framing; 	all models shift in the predicted direction (expert~$>$~human).%(bootstrap 95\% CI [4.73, 7.35]; 
	
\begin{figure}[h]
		\centering
		\caption{Poisson $\hat{\tau}$ (CH) across opponent identities for each model.}
		
		\includegraphics[width=\linewidth]{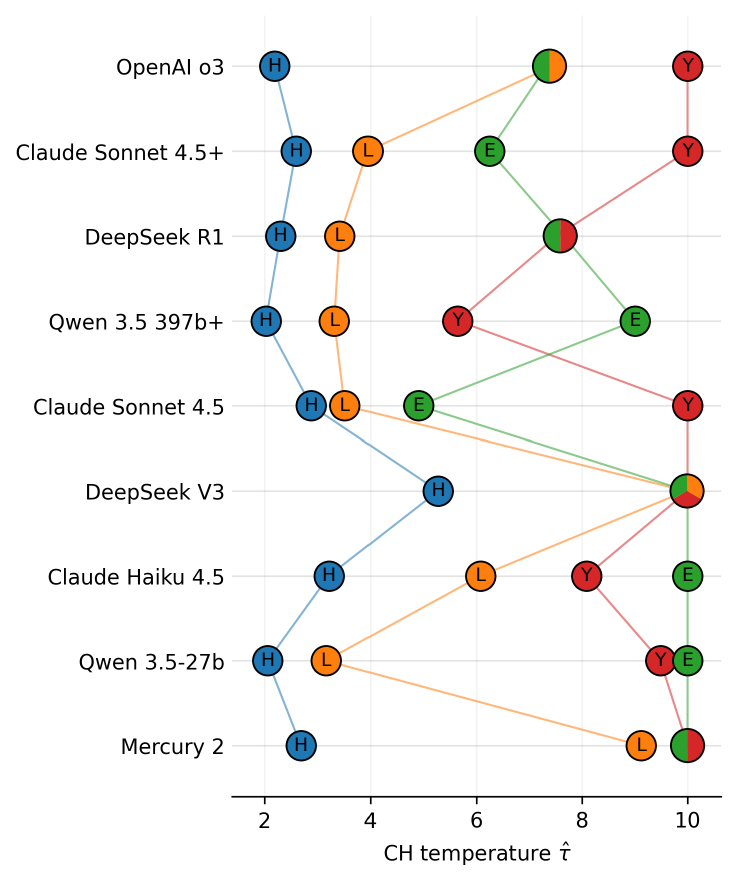}
		\label{fig:bcg-bias}
		\tabnotes{Each row corresponds to one model, with colored circles indicating opponent types\\(H = Human, L = LLM, E = Expert, Y = Yourself).}
	\end{figure}

	This analysis reveals interesting insights into meta-thinking processes of LLMs. In particular, we observe that not all models assign the highest possible level of depth of thinking to "unspecified" LLMs, motivating this conjecture by the fact that these models are trained on human data. For instance, some models reason that because an LLM is trained on human generated outputs, they would probably play one level higher than them, which in turn implies that it should play 1-2 levels higher than a human (Figure \ref{fig:bcg-bias} and Table~\ref{tab:heuristics_crossgame}, Appendix). 

	\subsection*{Strategic complexity induces shifts in reasoning logic}

	So far, we have seen that when we exogenously target specific levels of reasoning depth, models display coherent iterative best-response behavior: their choices maximize expected payoffs given the conjectured opponent behavior. However, once these exogenous constraints are removed, a subset of models transitions from explicit best-response reasoning to an equilibrium-based logic.

	As strategic complexity increases, models no longer sustain explicit recursive best-response reasoning throughout. In simple, convergent settings such as the BCG, recursion remains tractable and produces coherent level~k chains (e.g., as in Table~\ref{tab:backtrack}). In the UMG, without prior knowledge, most models still converge on the highest-payoff region, detect cycles, and settle on a fixed point within the cycle; a behavior consistent with heuristic resolution under indeterminacy (Figure~\ref{fig:UMG-expert}).
	
	On the other hand, in more complex games like the MRG, where best-response cycles do not converge, models reorganize their reasoning around equilibrium-compatible behavior, often explicitly invoking solution concepts in the traces (Tables~\ref{tab:heuristics_crossgame},~\ref{tab:keywords}, Appendix) and concentrating choices within the equilibrium support (Figure~\ref{fig:MRG-comparison}). Others bypass equilibrium reasoning altogether and directly produce heuristic arguments, selecting focal or boundary actions without iterating beliefs.
	
	The observed choice patterns and reasoning traces invoking equilibrium or simplified heuristic criteria show that, under increasing complexity, LLMs do not simply deepen recursion indefinitely but reconfigure their reasoning logic: they either substitute equilibrium reasoning for iterated best-response or collapse into simple rule-based choice. This transition suggests a form of bounded recursion, in which the model’s reasoning architecture shifts from explicit iterative best-response modeling to simplified responses to strategic interdependence.

	\subsection*{Probabilistic models do not necessarily implement probabilistic strategies}
	
	The shift toward equilibrium-based reasoning is most clearly illustrated in the MRG task, which admits no pure-strategy Nash equilibrium, so iterated best-response does not converge in pure strategies. In repeated experimental evaluations, no evidence of such cyclical play is observed. Instead, several models consistently select actions lying within the support of the (unique) mixed-strategy Nash equilibrium, particularly when prompted to play against experts (Figure~\ref{fig:MRG-comparison}). 
	
	Yet, even when this shift in logic occurs, as evidenced both through choices and traces, LLMs do not reproduce the theoretical distribution of choice predicted by the mixed Nash equilibrium of the game. Across repeated runs, action frequencies remain narrowly concentrated on a single option or a small subset of actions within the support of the equilibrium. 
	
	The architectural stochasticity of LLMs, sampling tokens from conditional probability distributions, does not propagate to the level of strategic choice. Instead, models approximate probabilistic equilibrium behavior somewhat deterministically, effectively collapsing the equilibrium distribution onto one or a few focal points rather than operationalizing strategic indifference through probabilistic choice. Increasing the sampling temperature (from $t=0.25$ to $t=0.75$), thereby amplifying stochasticity at inference time, does not meaningfully alter this pattern. Hence, despite their probabilistic architecture, LLMs behave as deterministic implementers in contexts where equilibrium play theoretically requires probabilistic strategists.

	\subsection*{Heuristic reasoning emerges under complexity and indeterminacy}
	
	When recursive reasoning fails to converge, models resolve indeterminacy through stable heuristic rules to select actions. For models that do not transition to equilibrium logic, heuristics serve as a means of approximating best responses rather than replacing them. In these cases, models apply simple structural rules such as focusing on boundary or symmetry points to identify what they treat as payoff-maximizing actions, without performing an iterated simulation of opponents' behavior. For models that adopt equilibrium reasoning, heuristics operate within that framework. When equilibrium play requires probabilistic mixing, as in the MRG, models do not implement the implied distributions. Instead, they deterministically select one equilibrium-consistent action, typically guided by focal or symmetry-based cues (see Table~\ref{tab:heuristics_crossgame}, Appendix for examples of rationales). 
	
	Across plays, the heuristics used within and outside equilibrium reasoning differ. The use of a given heuristic in one logic does not imply its use in the other. For example, models that select the lower bound of the feasible set when applying best-response logic do not necessarily select the upper bound of the equilibrium support in the MRG. Yet choice behavior consistently gravitates toward focal points such as upper and lower bounds of intervals, suggesting a form of structural salience guiding model behavior under strategic complexity.

	A possible explanation is that structural salience arises when models sustain deeper recursive simulation before emitting the first token. One hypothesis, consistent with reports of pre-output planning in larger models~\cite{_TracingThoughtsLarge_}, is that capacity to sustain latent recursion grows with scale. As shown in Figure~\ref{fig:MRG-emergence-expert}, while all models concentrate on salient extremes (actions 11, 19, 20), focalization around the mixed-equilibrium support (actions 15-20) appears in the Qwen family and grows more distributed across the support in the larger Qwen~3.5 397b+ variant, consistent with the scale-dependent focalization pattern reported above\footnote{This finding replicates to Google's Gemini family of models which we added here as a robustness check.}. This view aligns with Anthropic's findings that models plan ahead internally and that reported chains-of-thought are not complete records of reasoning.\footnote{The observation that substantial computation occurs before token emission, and therefore outside the verbalized reasoning trace, underscores the need for a hybrid analytical approach. By combining reasoning traces with revealed choices, we can infer reasoning logics even when the verbal record is incomplete. In this sense, choice data provide behavioral evidence that complements partial introspective evidence from text.} It also fits our earlier result that token count during inference does not predict best-response accuracy: reasoning length in text reflects verbalization, not the depth of internal computation. 
	
	Figure~\ref{fig:heuristic-text} shows that the model's reasoning vocabulary shifts with opponent identity: facing a game theory expert triggers Nash-equilibrium reasoning (mixed strategies, indifference conditions, cyclic best-response structure), while facing a human triggers level-$k$ iteration and maximin heuristics. This vocabulary shift, jointly with the choice shift in Fig.~\ref{fig:bcg-bias}, indicates that the heuristics are expressed in both reasoning traces and revealed choices rather than arising from stochastic noise.

	Taken together, these findings indicate that heuristic reasoning emerges as a dominant adaptive mechanism through which large language models stabilize their choice behavior under strategic complexity and indeterminacy. Unlike classic human biases such as anchoring or framing, these patterns illustrate novel \textit{emergent heuristics}: reasoning shortcuts or proper simplified rules that LLMs invent for a task, distinct from human cognitive biases or equilibrium concepts. In addition, models that perform similarly on traditional benchmarks can diverge in their use of these heuristics, suggesting that they are artifacts of training rather than inevitable consequences of the transformer architecture itself.

	\section{Discussion and Conclusions}

	\subsection*{Summary of Contributions}
	
	\textit{Can LLMs think strategically?} We find that LLMs can engage in the three key components of strategic thinking, that is, forming conjectures about other players, evaluating options, and choosing best responses, but that this capability is fragile under increased strategic complexity, even in simple noiseless environments. To demonstrate this, we used a hybrid method combining choice behavior and reasoning trace across targeted and unconstrained reasoning depths. This approach allows us to infer whether LLMs exhibit consistent conjectures about agents when placed in interactive situations, whether they can correctly evaluate the conjectures' strategic implications, and whether they can optimally act upon them. As it relates to the antecedents of inconsistent behavior in this domain, we further observe that most deviations from best-response behavior can be ascribed to heterogeneous limitations in emergent computational abilities of different LLMs, and not to computational timeout. 
		We also observe that different models vary in their degree of overthinking, defined as differences in efficiency of paths to final response. However, whether this type of overthinking is a feature more than a bug, and thus justifies the higher computational costs incurred for the process, remains unexplored, as this would relate to a possible further emergent LLM ability to question and verify their own responses across different contexts.

	Core to our contribution is the development of a structured method to measure strategic thinking, and we ensure that our findings do not derive from training data memorization (Appendix~\ref{sec:appendix-knowledge}).
	To this aim, we designed the UMG with a novel payoff matrix, so that the specific equilibrium and any solution-specific reasoning cannot be retrieved from training data, thereby isolating in-context computation from cached solutions. Through a prior knowledge check we find that the three games cover the spectrum of completely unknown (UMG) to partly known (MRG) and fully known (BCG). The convergence of results across all three games therefore supports the interpretation that best-response behavior reflects in-context strategic reasoning rather than retrieved knowledge.
	
	We further find that, as complexity increases, LLMs develop internally motivated heuristics, suggesting that strategic cognition can emerge from language modeling objectives in ways that differ from boundaries on deductive human reasoning.

This behavior in complex strategic interactions is further reminiscent of a form of \textit{pseudo-recall} of possibly related solution concepts \cite{kalai2025language}. In particular, when iterative best-response reasoning does not converge in strategic interactions where an equilibrium nevertheless exists, LLMs will sometimes shift from a best response logic to an equilibrium one in choice behavior. 

	Contextually, whenever we observe LLMs shifting to a probabilistic equilibrium logic, instead of replicating equilibrium choice distributions, they consistently adopt heuristic-driven selection mechanisms. This implies that LLMs react to strategic complexity in different, non-mutually exclusive, ways. 
	These shifts, however, are not mere computational timeouts, nor are they direct implications of strategic reasoning. They are model-specific procedural adaptations occurring under increased strategic complexity, even in controlled, noiseless, and relatively simple environments.
	This finding has substantial implications. Strategic thinking hinges on the possibility of putting oneself in the position of another player, forming a conjecture about that player's behavior, and deriving the behavioral implications of that conjecture. If this recursive structure becomes unpredictable under complexity, then the model's ability to form coherent conjectures is itself impaired. Since best-response behavior depends on those conjectures, the failure is not localized at the level of final choice; it propagates through the entire structure of strategic decision making. The fact that these deviations appear under controlled experimental conditions suggests that they may be more consequential in the unstructured, high-stakes environments in which agentic systems are increasingly being deployed.
    The model-specific nature of these logic shifts further exacerbates this concern. At any given point in time, different models may resolve strategic complexity through different procedures. An agent facing another synthetic agent may therefore be uncertain not only about the opponent's intended action, but also about the reasoning logic through which that action is generated. This weakens the basis for forming conjectures and deriving their implications. The problem is compounded when the nature of the opponent is itself uncertain: even if one could characterize the reasoning procedure of a given model, an agent may not know whether it is facing that model, another model, a human, or a hybrid decision process. Strategic prediction then becomes uncertain at two levels: the identity of the opponent and the logic by which that opponent maps the strategic environment into behavior. As models evolve over time, this unpredictability is likely to deepen rather than disappear.

	Our results suggest that some LLMs form opponent-specific conjectures by reasoning about training-data regularities, rather than only from the formal structure of the game.
	This is evidenced by the fact that LLMs motivate conjectures about synthetic agents through meta-thinking about training data and its implications for the opponent's likely behavior.
	Conditional on being able to form such structured conjectures about other agents, best-response behavior follows as a corollary of economic rationality~\cite{chen_EmergenceEconomicRationality_2023}.
	In our game-theoretic working definition, that is, maximization of expected payoff conditional on conjectures about other players, strategic thinking in LLMs therefore hinges on the ability of LLMs to derive those conjectures.

    If LLMs are used in high-stakes multi-agent negotiations, they must adapt their behavior to the other players. Our results show that, indeed, LLMs form differentiated conjectures about each player, assigning distinct levels of reasoning depth to humans and synthetic agents. This enriches the prior evidence of economically rational behavior (i.e. maximization of expected payoff) when LLMs are instructed to act as human decision makers~\cite{chen_EmergenceEconomicRationality_2023}, and highlights that strategic cognition in LLMs is not monolithic but opponent-contingent. It is therefore critical for future research to identify and study the component of strategic thinking that relates to \textit{how} synthetic agents form conjectures about their opponents' play.
	
    This work further highlights the currently open question as to whether language alone, in both its verbal and mathematical form, is a \textit{sufficient} representation for strategic thinking. In other words, does language encode all the information necessary to enact strategic thinking, or at least to behave \textit{as if} it were taking place? While in the context of this paper we restrict our attention to well-defined mathematical problems, the reach of the \textit{conditional best-response logic} of strategic thinking is broader and is routinely brought to bear on settings where the elements that constitute the game are more loosely defined and thus only approximate implications, rather than sharp predictions, can be derived. 
	From this perspective, our question is tied to whether language models can abstract the underlying dynamics %logic?
	of strategic reasoning from pretraining and transfer it by analogy to different, less well-defined applied contexts.
	
	\subsection*{Limitations}
	As a study of the complex phenomenon of emergence of strategic thinking ability in LLMs, this study has several limitations, opening the way to further future research. We did not do a deep-dive into the differences in dynamics that may arise when a strategic situation moves from two to a finite multiplicity of agents. Preliminary evidence suggests that GPTs may exhibit a shift towards lower depth of reasoning as the number of players increases. Yet, whether this behavior is robustly observed between different interactive choice situations is yet to be studied. Relatedly, the effect of context, intended as the specific strategic situation under analysis, on LLMs' beliefs about their opponents’ behavior remains to be systematically examined. Empirical research in experimental and behavioral game theory shows that human reasoning depth and belief formation are not consistent across games~\cite{georganas_PersistenceStrategicSophistication_2015a}. Whether LLMs display similar context-dependent variation in their inferred reasoning depth therefore represents an important question for future research. 
	In addition, our hybrid method, when looking at consistency of trace by keywords and choice behavior, to obtain additional insights in addition to those emerging from revealed preference behavior, relies on faithfulness of Chain-of-Thought~\cite{chen_ReasoningModelsDont_2025}. In our experiments, we observe only a mild, yet consistent, impact of tracing on implied depth of reasoning, thus mitigating this concern.
	
	\subsection*{Implications and future work}
	Our study shows that strategic thinking can emerge in surprising ways in LLMs in structured contexts of varying complexity, where objectives are formally specified. Future research is needed to analyze this ability and emerging behavior when situations with strategic interdependencies are unstructured and lack formal instructions on the gameplay.

	While existing reasoning benchmarks are a useful tool, a large subset of real-life agentic applications with reasoning at their core typically require recursive belief modeling and formulation of priors that are not covered by these types of evaluations~\cite{shojaee_IllusionThinkingUnderstanding_2025}. It is therefore no surprise that evidence is emerging that LLMs frequently exhibit systematic errors in strategic contexts, particularly in dynamic multi-agent scenarios~\cite{duan_BotChatEvaluatingLLMs_2024,wang_EmergentHierarchicalReasoning_2025,horton_LargeLanguageModels_2023,junquedefortuny_SimulatingMarketEquilibrium_2025}. There is a pressing need for more rigorous theoretical frameworks to evaluate the reasoning capabilities of LLMs in strategically interdependent settings. 

	As in social computing more broadly, these frameworks cannot rely on formal task structure alone: they must account for the behavior of the agents embedded in the system, since deviations from standard agent models can alter system-level conclusions. Once such capabilities are reliably verified, measurable progress toward true strategic reasoning will become more likely~\cite{wei_AsymmetryVerificationVerifiers_2025}. Developing a structured method to observe this capability in noiseless setups represents only the first step towards understanding the reach of this instance of emergent abilities in LLM and to monitor their evolution.

	While our testing protocol is theoretical, the relevance of the question extends beyond controlled game-theoretic environments. LLM-based systems are already deployed as autonomous strategic actors in consequential commercial settings: AI agents conduct supplier negotiations at scale in the retail and logistics industries~\cite{pactum2024}, and LLM-informed systems guide trading decisions at major hedge funds under conditions of limited regulatory oversight~\cite{hsgac2024}. Financial regulators are actively monitoring these deployments and their systemic implications~\cite{fed2025,bis2025}. In all of these settings, the distinction between genuine best-response reasoning and equilibrium-consistent heuristic behavior is immediately consequential: if similar strategic behavior can arise from different underlying reasoning processes or implied assumptions, then prompt design, steering, and oversight cannot rely on observed behavior alone.

Finally, this work feeds back into a reinvestigation of the study of normative behavior of language models. Decision science has traditionally looked at the implications (equivalence) of desirable properties of the decision maker on observable choice behavior and criteria. However, this raises the question about whether or to which extent we should apply the same normative perspective on synthetic systems. This leads to a contemporaneous study and reinvestigation of the normative desirability or epistemic compellingness of traditional solution concepts in strategic settings, such as Nash equilibrium, and thus their potential for applicability in agentic decision making. The implied analysis would serve as guidance for the analysis of decision making in agentic systems, contraposing what should we normatively expect against what is descriptively observed, thus leading to a structured analysis of these potential deviations in synthetic decision making, which can inform differences in emergent "reasoning" dynamics. This direction opens a dialogue between computer and decision science, prompting each side to potentially question and revisit the foundations of the analysis of decision making in both synthetic and human systems.
	
	\section*{Data Availability}
	Code and data are available at~\cite{junquedefortuny2026strategicthinkingdata}.
	
	\bibliographystyle{ACM-Reference-Format}
	\bibliography{bibliography-zotero}


\begin{thebibliography}{42}

%%% the Web address.

\ifx \showCODEN    \undefined \def \showCODEN     #1{\unskip}     \fi
\ifx \showISBNx    \undefined \def \showISBNx     #1{\unskip}     \fi
\ifx \showISBNxiii \undefined \def \showISBNxiii  #1{\unskip}     \fi
\ifx \showISSN     \undefined \def \showISSN      #1{\unskip}     \fi
\ifx \showLCCN     \undefined \def \showLCCN      #1{\unskip}     \fi
\ifx \shownote     \undefined \def \shownote      #1{#1}          \fi
\ifx \showarticletitle \undefined \def \showarticletitle #1{#1}   \fi
\ifx \showURL      \undefined \def \showURL       {\relax}        \fi
% invisible to TeX
\providecommand\bibfield[2]{#2}
\providecommand\bibinfo[2]{#2}
\providecommand\natexlab[1]{#1}
\providecommand\showeprint[2][]{arXiv:#2}

\bibitem[_AI(2025)]%
        {_AIModelsCould_2025}
 \bibinfo{year}{2025}\natexlab{}.
\newblock \showarticletitle{{{AI}} Models Could Help Negotiators Secure Peace
  Deals}.
\newblock \bibinfo{journal}{\emph{The Economist}} (\bibinfo{year}{2025}).
\newblock
\showISSN{0013-0613}


\bibitem[Alekseenko et~al\mbox{.}(2025)]%
        {alekseenko2025strategizing}
\bibfield{author}{\bibinfo{person}{Iuliia Alekseenko}, \bibinfo{person}{Dmitry
  Dagaev}, \bibinfo{person}{Sofiia Paklina}, {and} \bibinfo{person}{Petr
  Parshakov}.} \bibinfo{year}{2025}\natexlab{}.
\newblock \showarticletitle{Strategizing with {{AI}}: {{Insights}} from a
  Beauty Contest Experiment}.
\newblock \bibinfo{journal}{\emph{Journal of Economic Behavior \&
  Organization}}  \bibinfo{volume}{240} (\bibinfo{year}{2025}),
  \bibinfo{pages}{107330}.
\newblock
\href{https://doi.org/10.1016/j.jebo.2025.107330}{doi:\nolinkurl{10.1016/j.jebo.2025.107330}}


\bibitem[{Anthropic}(2025)]%
        {_TracingThoughtsLarge_}
\bibfield{author}{\bibinfo{person}{{Anthropic}}.}
  \bibinfo{year}{2025}\natexlab{}.
\newblock \bibinfo{title}{Tracing the Thoughts of a Large Language Model}.
\newblock
\urldef\tempurl%
\url{https://www.anthropic.com/news/tracing-thoughts-language-model}
\showURL{%
\tempurl}


\bibitem[Arad and Rubinstein(2012)]%
        {arad_1120MoneyRequest_2012}
\bibfield{author}{\bibinfo{person}{Ayala Arad} {and} \bibinfo{person}{Ariel
  Rubinstein}.} \bibinfo{year}{2012}\natexlab{}.
\newblock \showarticletitle{The 11-20 {{Money Request Game}}: {{A Level-k
  Reasoning Study}}}.
\newblock \bibinfo{journal}{\emph{American Economic Review}}
  \bibinfo{volume}{102}, \bibinfo{number}{7} (\bibinfo{date}{Dec.}
  \bibinfo{year}{2012}), \bibinfo{pages}{3561--3573}.
\newblock
\showISSN{0002-8282}
\href{https://doi.org/10.1257/aer.102.7.3561}{doi:\nolinkurl{10.1257/aer.102.7.3561}}


\bibitem[Battigalli et~al\mbox{.}(2023)]%
        {battigalli_GAMETHEORYAnalysis_2023}
\bibfield{author}{\bibinfo{person}{Pierpaolo Battigalli},
  \bibinfo{person}{Emiliano Catonini}, {and} \bibinfo{person}{Nicodemo~De
  Vito}.} \bibinfo{year}{2023}\natexlab{}.
\newblock \bibinfo{booktitle}{\emph{{{GAME THEORY}}: {{Analysis}} of
  {{Strategic Thinking}}}}.
\newblock


\bibitem[Bini et~al\mbox{.}(2026)]%
        {bini2026behavioral}
\bibfield{author}{\bibinfo{person}{Pietro Bini}, \bibinfo{person}{Lin~William
  Cong}, \bibinfo{person}{Xing Huang}, {and} \bibinfo{person}{Lawrence~J.
  Jin}.} \bibinfo{year}{2026}\natexlab{}.
\newblock \bibinfo{booktitle}{\emph{Behavioral Economics of {{AI}}: {{LLM}}
  Biases and Corrections}}.
\newblock \bibinfo{type}{Working {{Paper}}} 34745.
  \bibinfo{institution}{National Bureau of Economic Research}.
\newblock
\href{https://doi.org/10.3386/w34745}{doi:\nolinkurl{10.3386/w34745}}


\bibitem[{Board of Governors of the Federal Reserve System}(2025)]%
        {fed2025}
\bibfield{author}{\bibinfo{person}{{Board of Governors of the Federal Reserve
  System}}.} \bibinfo{year}{2025}\natexlab{}.
\newblock \bibinfo{booktitle}{\emph{Financial Stability Implications of
  Generative {{AI}}}}.
\newblock \bibinfo{type}{{T}echnical {R}eport} 2025-090.
  \bibinfo{institution}{Federal Reserve}.
\newblock


\bibitem[Chen et~al\mbox{.}(2025)]%
        {chen_ReasoningModelsDont_2025}
\bibfield{author}{\bibinfo{person}{Yanda Chen}, \bibinfo{person}{Joe Benton},
  \bibinfo{person}{Ansh Radhakrishnan}, \bibinfo{person}{Jonathan Uesato},
  \bibinfo{person}{Carson Denison}, \bibinfo{person}{John Schulman},
  \bibinfo{person}{Arushi Somani}, \bibinfo{person}{Peter Hase},
  \bibinfo{person}{Misha Wagner}, \bibinfo{person}{Fabien Roger},
  \bibinfo{person}{Vlad Mikulik}, \bibinfo{person}{Sam Bowman},
  \bibinfo{person}{Jan Leike}, \bibinfo{person}{Jared Kaplan}, {and}
  \bibinfo{person}{Ethan Perez}.} \bibinfo{year}{2025}\natexlab{}.
\newblock \showarticletitle{Reasoning {{Models Don}}'t {{Always Say What They
  Think}}}.
\newblock  (\bibinfo{year}{2025}).
\newblock


\bibitem[Chen et~al\mbox{.}(2023)]%
        {chen_EmergenceEconomicRationality_2023}
\bibfield{author}{\bibinfo{person}{Yiting Chen}, \bibinfo{person}{Tracy~Xiao
  Liu}, \bibinfo{person}{You Shan}, {and} \bibinfo{person}{Songfa Zhong}.}
  \bibinfo{year}{2023}\natexlab{}.
\newblock \showarticletitle{The Emergence of Economic Rationality of {{GPT}}}.
\newblock \bibinfo{journal}{\emph{Proceedings of the National Academy of
  Sciences}} \bibinfo{volume}{120}, \bibinfo{number}{51} (\bibinfo{date}{Dec.}
  \bibinfo{year}{2023}), \bibinfo{pages}{e2316205120}.
\newblock
\showISSN{0027-8424, 1091-6490}
\href{https://doi.org/10.1073/pnas.2316205120}{doi:\nolinkurl{10.1073/pnas.2316205120}}


\bibitem[Coz et~al\mbox{.}(2025)]%
        {coz_WhatWouldLLM_2025}
\bibfield{author}{\bibinfo{person}{Pierre~Le Coz}, \bibinfo{person}{Jia~An
  Liu}, \bibinfo{person}{Debarun Bhattacharjya}, \bibinfo{person}{Georgina
  Curto}, {and} \bibinfo{person}{Serge Stinckwich}.}
  \bibinfo{year}{2025}\natexlab{}.
\newblock \bibinfo{title}{What {{Would}} an {{LLM Do}}? {{Evaluating
  Policymaking Capabilities}} of {{Large Language Models}}}.
\newblock
\href{https://doi.org/10.48550/arXiv.2509.03827}{doi:\nolinkurl{10.48550/arXiv.2509.03827}}
\showeprint[arxiv]{2509.03827}~[cs]


\bibitem[{DeepSeek-AI}(2025)]%
        {deepseekr1_2025}
\bibfield{author}{\bibinfo{person}{{DeepSeek-AI}}.}
  \bibinfo{year}{2025}\natexlab{}.
\newblock \showarticletitle{{{DeepSeek-R1}}: {{Incentivizing}} Reasoning
  Capability in {{LLMs}} via Reinforcement Learning}.
\newblock \bibinfo{journal}{\emph{arXiv preprint arXiv:2501.12948}}
  (\bibinfo{year}{2025}).
\newblock
\href{https://doi.org/10.1038/s41586-025-09422-z}{doi:\nolinkurl{10.1038/s41586-025-09422-z}}
\showeprint[arxiv]{2501.12948}


\bibitem[Duan et~al\mbox{.}(2024a)]%
        {duan_BotChatEvaluatingLLMs_2024}
\bibfield{author}{\bibinfo{person}{Haodong Duan}, \bibinfo{person}{Jueqi Wei},
  \bibinfo{person}{Chonghua Wang}, \bibinfo{person}{Hongwei Liu},
  \bibinfo{person}{Yixiao Fang}, \bibinfo{person}{Songyang Zhang},
  \bibinfo{person}{Dahua Lin}, {and} \bibinfo{person}{Kai Chen}.}
  \bibinfo{year}{2024}\natexlab{a}.
\newblock \showarticletitle{{{BotChat}}: {{Evaluating LLMs}}' {{Capabilities}}
  of {{Having Multi-Turn Dialogues}}}. In \bibinfo{booktitle}{\emph{Findings of
  the {{Association}} for {{Computational Linguistics}}: {{NAACL}} 2024}},
  \bibfield{editor}{\bibinfo{person}{Kevin Duh}, \bibinfo{person}{Helena
  Gomez}, {and} \bibinfo{person}{Steven Bethard}} (Eds.).
  \bibinfo{publisher}{Association for Computational Linguistics},
  \bibinfo{address}{Mexico City, Mexico}, \bibinfo{pages}{3184--3200}.
\newblock
\href{https://doi.org/10.18653/v1/2024.findings-naacl.201}{doi:\nolinkurl{10.18653/v1/2024.findings-naacl.201}}


\bibitem[Duan et~al\mbox{.}(2024b)]%
        {duan_GTBenchUncoveringStrategic_2024}
\bibfield{author}{\bibinfo{person}{Jinhao Duan}, \bibinfo{person}{Renming
  Zhang}, \bibinfo{person}{James Diffenderfer}, \bibinfo{person}{Bhavya
  Kailkhura}, \bibinfo{person}{Lichao Sun}, \bibinfo{person}{Elias
  {Stengel-Eskin}}, \bibinfo{person}{Mohit Bansal}, \bibinfo{person}{Tianlong
  Chen}, {and} \bibinfo{person}{Kaidi Xu}.} \bibinfo{year}{2024}\natexlab{b}.
\newblock \bibinfo{title}{{{GTBench}}: {{Uncovering}} the {{Strategic Reasoning
  Limitations}} of {{LLMs}} via {{Game-Theoretic Evaluations}}}.
\newblock
\href{https://doi.org/10.48550/arXiv.2402.12348}{doi:\nolinkurl{10.48550/arXiv.2402.12348}}
\showeprint[arxiv]{2402.12348}~[cs]


\bibitem[Echterhoff et~al\mbox{.}(2024)]%
        {echterhoff_CognitiveBiasDecisionMaking_2024}
\bibfield{author}{\bibinfo{person}{Jessica Echterhoff}, \bibinfo{person}{Yao
  Liu}, \bibinfo{person}{Abeer Alessa}, \bibinfo{person}{Julian McAuley}, {and}
  \bibinfo{person}{Zexue He}.} \bibinfo{year}{2024}\natexlab{}.
\newblock \bibinfo{title}{Cognitive {{Bias}} in {{Decision-Making}} with
  {{LLMs}}}.
\newblock
\href{https://doi.org/10.48550/arXiv.2403.00811}{doi:\nolinkurl{10.48550/arXiv.2403.00811}}
\showeprint[arxiv]{2403.00811}~[cs]


\bibitem[Freund et~al\mbox{.}(1995)]%
        {freund_EfficientAlgorithmsLearning_1995}
\bibfield{author}{\bibinfo{person}{Y. Freund}, \bibinfo{person}{M. Kearns},
  \bibinfo{person}{Y. Mansour}, \bibinfo{person}{D. Ron}, \bibinfo{person}{R.
  Rubinfeld}, {and} \bibinfo{person}{R.E. Schapire}.}
  \bibinfo{year}{1995}\natexlab{}.
\newblock \showarticletitle{Efficient Algorithms for Learning to Play Repeated
  Games against Computationally Bounded Adversaries}. In
  \bibinfo{booktitle}{\emph{Proceedings of {{IEEE}} 36th {{Annual Foundations}}
  of {{Computer Science}}}}. \bibinfo{publisher}{IEEE Comput. Soc. Press},
  \bibinfo{address}{Milwaukee, WI, USA}, \bibinfo{pages}{332--341}.
\newblock
\showISBNx{978-0-8186-7183-8}
\href{https://doi.org/10.1109/SFCS.1995.492489}{doi:\nolinkurl{10.1109/SFCS.1995.492489}}


\bibitem[Georganas et~al\mbox{.}(2015)]%
        {georganas_PersistenceStrategicSophistication_2015a}
\bibfield{author}{\bibinfo{person}{Sotiris Georganas}, \bibinfo{person}{Paul~J.
  Healy}, {and} \bibinfo{person}{Roberto~A. Weber}.}
  \bibinfo{year}{2015}\natexlab{}.
\newblock \showarticletitle{On the Persistence of Strategic Sophistication}.
\newblock \bibinfo{journal}{\emph{Journal of Economic Theory}}
  \bibinfo{volume}{159} (\bibinfo{date}{Sept.} \bibinfo{year}{2015}),
  \bibinfo{pages}{369--400}.
\newblock
\showISSN{00220531}
\href{https://doi.org/10.1016/j.jet.2015.07.012}{doi:\nolinkurl{10.1016/j.jet.2015.07.012}}


\bibitem[Horton(2023)]%
        {horton_LargeLanguageModels_2023}
\bibfield{author}{\bibinfo{person}{John~J. Horton}.}
  \bibinfo{year}{2023}\natexlab{}.
\newblock \bibinfo{title}{Large {{Language Models}} as {{Simulated Economic
  Agents}}: {{What Can We Learn}} from {{Homo Silicus}}?}
\newblock
\href{https://doi.org/10.48550/arXiv.2301.07543}{doi:\nolinkurl{10.48550/arXiv.2301.07543}}
\showeprint[arxiv]{2301.07543}~[econ]


\bibitem[Junqu{\'e} De~Fortuny(2025)]%
        {junquedefortuny_SimulatingMarketEquilibrium_2025}
\bibfield{author}{\bibinfo{person}{Enric Junqu{\'e} De~Fortuny}.}
  \bibinfo{year}{2025}\natexlab{}.
\newblock \showarticletitle{Simulating {{Market Equilibrium}} with {{Large
  Language Models}}}. In \bibinfo{booktitle}{\emph{Hawaii {{International
  Conference}} on {{System Sciences}}}}.
\newblock
\href{https://doi.org/10.24251/HICSS.2025.599}{doi:\nolinkurl{10.24251/HICSS.2025.599}}
\showeprint[hdl]{10125/109446}


\bibitem[{Junqu{\'e} de Fortuny} and Cappelli(2026)]%
        {junquedefortuny2026strategicthinkingdata}
\bibfield{author}{\bibinfo{person}{Enric {Junqu{\'e} de Fortuny}} {and}
  \bibinfo{person}{Veronica~Roberta Cappelli}.}
  \bibinfo{year}{2026}\natexlab{}.
\newblock \bibinfo{title}{Data and Code for: {{The}} Fragility of Strategic
  Thinking in Large Language Models}.
\newblock
\href{https://doi.org/TO-BE-ASSIGNED}{doi:\nolinkurl{TO-BE-ASSIGNED}}


\bibitem[Kalai et~al\mbox{.}(2025)]%
        {kalai2025language}
\bibfield{author}{\bibinfo{person}{Adam~Tauman Kalai}, \bibinfo{person}{Ofir
  Nachum}, \bibinfo{person}{Santosh~S Vempala}, {and} \bibinfo{person}{Edwin
  Zhang}.} \bibinfo{year}{2025}\natexlab{}.
\newblock \showarticletitle{Why Language Models Hallucinate}.
\newblock \bibinfo{journal}{\emph{arXiv preprint arXiv:2509.04664}}
  (\bibinfo{year}{2025}).
\newblock
\showeprint[arxiv]{2509.04664}


\bibitem[Kempinski et~al\mbox{.}(2025)]%
        {kempinski2025game}
\bibfield{author}{\bibinfo{person}{Benjamin Kempinski}, \bibinfo{person}{Ian
  Gemp}, \bibinfo{person}{Kate Larson}, \bibinfo{person}{Marc Lanctot},
  \bibinfo{person}{Yoram Bachrach}, {and} \bibinfo{person}{Tal Kachman}.}
  \bibinfo{year}{2025}\natexlab{}.
\newblock \showarticletitle{Game of Thoughts: {{Iterative}} Reasoning in
  Game-Theoretic Domains with Large Language Models}.
\newblock  (\bibinfo{year}{2025}).
\newblock
\href{https://doi.org/10.65109/gzfu8152}{doi:\nolinkurl{10.65109/gzfu8152}}


\bibitem[Lee and Kader(2025)]%
        {lee_EmergenceStrategicReasoning_2025}
\bibfield{author}{\bibinfo{person}{Dongwoo Lee} {and} \bibinfo{person}{Gavin
  Kader}.} \bibinfo{year}{2025}\natexlab{}.
\newblock \bibinfo{title}{The {{Emergence}} of {{Strategic Reasoning}} of
  {{Large Language Models}}}.
\newblock
\href{https://doi.org/10.48550/arXiv.2412.13013v3}{doi:\nolinkurl{10.48550/arXiv.2412.13013v3}}
\showeprint[arxiv]{2412.13013}~[econ]


\bibitem[Lor{\`e} and Heydari(2024)]%
        {lore_StrategicBehaviorLarge_2024}
\bibfield{author}{\bibinfo{person}{Nunzio Lor{\`e}} {and}
  \bibinfo{person}{Babak Heydari}.} \bibinfo{year}{2024}\natexlab{}.
\newblock \showarticletitle{Strategic Behavior of Large Language Models and the
  Role of Game Structure versus Contextual Framing}.
\newblock \bibinfo{journal}{\emph{Scientific Reports}} \bibinfo{volume}{14},
  \bibinfo{number}{1} (\bibinfo{date}{Aug.} \bibinfo{year}{2024}),
  \bibinfo{pages}{18490}.
\newblock
\showISSN{2045-2322}
\href{https://doi.org/10.1038/s41598-024-69032-z}{doi:\nolinkurl{10.1038/s41598-024-69032-z}}


\bibitem[Lu(2025)]%
        {lu2025beauty}
\bibfield{author}{\bibinfo{person}{Siting~Estee Lu}.}
  \bibinfo{year}{2025}\natexlab{}.
\newblock \showarticletitle{Game-Theory Behaviour of Large Language Models:
  {{The}} Case of {{Keynesian}} Beauty Contests}.
\newblock \bibinfo{journal}{\emph{Economics and Business Review}}
  \bibinfo{volume}{11}, \bibinfo{number}{2} (\bibinfo{year}{2025}),
  \bibinfo{pages}{119--148}.
\newblock
\href{https://doi.org/10.18559/ebr.2025.2.2182}{doi:\nolinkurl{10.18559/ebr.2025.2.2182}}


\bibitem[{Macmillan-Scott} and Musolesi(2024)]%
        {macmillanscott2024irrationality}
\bibfield{author}{\bibinfo{person}{Olivia {Macmillan-Scott}} {and}
  \bibinfo{person}{Mirco Musolesi}.} \bibinfo{year}{2024}\natexlab{}.
\newblock \showarticletitle{({{Ir}})Rationality and Cognitive Biases in Large
  Language Models}.
\newblock \bibinfo{journal}{\emph{Royal Society Open Science}}
  \bibinfo{volume}{11}, \bibinfo{number}{6} (\bibinfo{year}{2024}),
  \bibinfo{pages}{240255}.
\newblock
\href{https://doi.org/10.1098/rsos.240255}{doi:\nolinkurl{10.1098/rsos.240255}}


\bibitem[Manning and Horton(2025)]%
        {manning_GeneralSocialAgents_2025}
\bibfield{author}{\bibinfo{person}{Benjamin~S. Manning} {and}
  \bibinfo{person}{John~J. Horton}.} \bibinfo{year}{2025}\natexlab{}.
\newblock \bibinfo{title}{General {{Social Agents}}}.
\newblock
\href{https://doi.org/10.48550/arXiv.2508.17407v4}{doi:\nolinkurl{10.48550/arXiv.2508.17407v4}}
\showeprint[arxiv]{2508.17407v4}~[econ]


\bibitem[{Matteo Aquilina} et~al\mbox{.}(2025)]%
        {bis2025}
\bibfield{author}{\bibinfo{person}{{Matteo Aquilina}}, \bibinfo{person}{Douglas
  Araujo}, \bibinfo{person}{Gaston Gelos}, \bibinfo{person}{Taejin Park}, {and}
  \bibinfo{person}{Fernando {P{\'e}rez-Cruz}}.}
  \bibinfo{year}{2025}\natexlab{}.
\newblock \bibinfo{booktitle}{\emph{Harnessing Artificial Intelligence for
  Monitoring Financial Markets, Not {{Artificial}} Intelligence in Financial
  Markets}}.
\newblock \bibinfo{type}{{T}echnical {R}eport} BIS Working Papers 1291.
  \bibinfo{institution}{Bank for International Settlements}.
\newblock
\href{https://doi.org/10.1109/ICNN.1995.487527}{doi:\nolinkurl{10.1109/ICNN.1995.487527}}


\bibitem[Nagel(1995)]%
        {nagel_UnravelingGuessingGames_1995}
\bibfield{author}{\bibinfo{person}{Rosemarie Nagel}.}
  \bibinfo{year}{1995}\natexlab{}.
\newblock \showarticletitle{Unraveling in {{Guessing Games}}: {{An Experimental
  Study}}}.
\newblock \bibinfo{journal}{\emph{The American Economic Review}}
  \bibinfo{volume}{85}, \bibinfo{number}{5} (\bibinfo{year}{1995}),
  \bibinfo{pages}{1313--1326}.
\newblock
\showISSN{0002-8282}
\showeprint[jstor]{2950991}


\bibitem[Narendra et~al\mbox{.}(2024)]%
        {narendra_EnhancingContractNegotiations_2024}
\bibfield{author}{\bibinfo{person}{Savinay Narendra}, \bibinfo{person}{Kaushal
  Shetty}, {and} \bibinfo{person}{Adwait Ratnaparkhi}.}
  \bibinfo{year}{2024}\natexlab{}.
\newblock \showarticletitle{Enhancing {{Contract Negotiations}} with
  {{LLM-Based Legal Document Comparison}}}. In
  \bibinfo{booktitle}{\emph{Proceedings of the {{Natural Legal Language
  Processing Workshop}} 2024}}, \bibfield{editor}{\bibinfo{person}{Nikolaos
  Aletras}, \bibinfo{person}{Ilias Chalkidis}, \bibinfo{person}{Leslie
  Barrett}, \bibinfo{person}{C{\u a}t{\u a}lina Goan{\textcommabelow t}{\u a}},
  \bibinfo{person}{Daniel {Preo{\textcommabelow t}iuc-Pietro}}, {and}
  \bibinfo{person}{Gerasimos Spanakis}} (Eds.). \bibinfo{publisher}{Association
  for Computational Linguistics}, \bibinfo{address}{Miami, FL, USA},
  \bibinfo{pages}{143--153}.
\newblock
\href{https://doi.org/10.18653/v1/2024.nllp-1.11}{doi:\nolinkurl{10.18653/v1/2024.nllp-1.11}}


\bibitem[Saeedi et~al\mbox{.}(2025)]%
        {saeedi_HeuristicsBiasesAI_2025}
\bibfield{author}{\bibinfo{person}{Payam Saeedi}, \bibinfo{person}{Mahsa
  Goodarzi}, {and} \bibinfo{person}{M.~Abdullah Canbaz}.}
  \bibinfo{year}{2025}\natexlab{}.
\newblock \showarticletitle{Heuristics and {{Biases}} in {{AI
  Decision-Making}}: {{Implications}} for {{Responsible AGI}}}. In
  \bibinfo{booktitle}{\emph{2025 6th {{International Conference}} on
  {{Artificial Intelligence}}, {{Robotics}} and {{Control}} ({{AIRC}})}}.
  \bibinfo{pages}{214--221}.
\newblock
\href{https://doi.org/10.1109/AIRC64931.2025.11077505}{doi:\nolinkurl{10.1109/AIRC64931.2025.11077505}}
\showeprint[arxiv]{2410.02820}~[cs]


\bibitem[Shojaee et~al\mbox{.}(2025)]%
        {shojaee_IllusionThinkingUnderstanding_2025}
\bibfield{author}{\bibinfo{person}{Parshin Shojaee}, \bibinfo{person}{Iman
  Mirzadeh}, \bibinfo{person}{Keivan Alizadeh}, \bibinfo{person}{Maxwell
  Horton}, \bibinfo{person}{Samy Bengio}, {and} \bibinfo{person}{Mehrdad
  Farajtabar}.} \bibinfo{year}{2025}\natexlab{}.
\newblock \bibinfo{title}{The {{Illusion}} of {{Thinking}}: {{Understanding}}
  the {{Strengths}} and {{Limitations}} of {{Reasoning Models}} via the
  {{Lens}} of {{Problem Complexity}}}.
\newblock
\href{https://doi.org/10.48550/arXiv.2506.06941v2}{doi:\nolinkurl{10.48550/arXiv.2506.06941v2}}
\showeprint[arxiv]{2506.06941v2}~[cs]


\bibitem[Suri et~al\mbox{.}(2023)]%
        {suri_LargeLanguageModels_2023}
\bibfield{author}{\bibinfo{person}{Gaurav Suri}, \bibinfo{person}{Lily~R.
  Slater}, \bibinfo{person}{Ali Ziaee}, {and} \bibinfo{person}{Morgan Nguyen}.}
  \bibinfo{year}{2023}\natexlab{}.
\newblock \bibinfo{title}{Do {{Large Language Models Show Decision Heuristics
  Similar}} to {{Humans}}? {{A Case Study Using GPT-3}}.5}.
\newblock
\href{https://doi.org/10.48550/arXiv.2305.04400}{doi:\nolinkurl{10.48550/arXiv.2305.04400}}
\showeprint[arxiv]{2305.04400}~[cs]


\bibitem[{U.S. Senate Homeland Security and Governmental Affairs
  Committee}(2024)]%
        {hsgac2024}
\bibfield{author}{\bibinfo{person}{{U.S. Senate Homeland Security and
  Governmental Affairs Committee}}.} \bibinfo{year}{2024}\natexlab{}.
\newblock \bibinfo{booktitle}{\emph{{{AI}} in the {{Real World}}: {{Hedge
  Funds}}' {{Use}} of {{Artificial Intelligence}} in {{Trading}}}}.
\newblock \bibinfo{type}{{T}echnical {R}eport}. \bibinfo{institution}{United
  States Senate}.
\newblock


\bibitem[Wang et~al\mbox{.}(2025)]%
        {wang_EmergentHierarchicalReasoning_2025}
\bibfield{author}{\bibinfo{person}{Haozhe Wang}, \bibinfo{person}{Qixin Xu},
  \bibinfo{person}{Che Liu}, \bibinfo{person}{Junhong Wu},
  \bibinfo{person}{Fangzhen Lin}, {and} \bibinfo{person}{Wenhu Chen}.}
  \bibinfo{year}{2025}\natexlab{}.
\newblock \bibinfo{title}{Emergent {{Hierarchical Reasoning}} in {{LLMs}}
  through {{Reinforcement Learning}}}.
\newblock
\href{https://doi.org/10.48550/arXiv.2509.03646}{doi:\nolinkurl{10.48550/arXiv.2509.03646}}
\showeprint[arxiv]{2509.03646}~[cs]


\bibitem[Wei(2025)]%
        {wei_AsymmetryVerificationVerifiers_2025}
\bibfield{author}{\bibinfo{person}{Jason Wei}.}
  \bibinfo{year}{2025}\natexlab{}.
\newblock \bibinfo{title}{Asymmetry of Verification and Verifier's Rule}.
\newblock


\bibitem[Wei et~al\mbox{.}(2022a)]%
        {wei_EmergentAbilitiesLarge_2022}
\bibfield{author}{\bibinfo{person}{Jason Wei}, \bibinfo{person}{Yi Tay},
  \bibinfo{person}{Rishi Bommasani}, \bibinfo{person}{Colin Raffel},
  \bibinfo{person}{Barret Zoph}, \bibinfo{person}{Sebastian Borgeaud},
  \bibinfo{person}{Dani Yogatama}, \bibinfo{person}{Maarten Bosma},
  \bibinfo{person}{Denny Zhou}, \bibinfo{person}{Donald Metzler},
  \bibinfo{person}{Ed~H Chi}, \bibinfo{person}{Tatsunori Hashimoto},
  \bibinfo{person}{Oriol Vinyals}, \bibinfo{person}{Percy Liang},
  \bibinfo{person}{Jeff Dean}, {and} \bibinfo{person}{William Fedus}.}
  \bibinfo{year}{2022}\natexlab{a}.
\newblock \showarticletitle{Emergent {{Abilities}} of {{Large Language
  Models}}}. In \bibinfo{booktitle}{\emph{Transactions on {{Machine Learning
  Research}}}}.
\newblock


\bibitem[Wei et~al\mbox{.}(2022b)]%
        {wei_ChainofThoughtPromptingElicits_2022}
\bibfield{author}{\bibinfo{person}{Jason Wei}, \bibinfo{person}{Xuezhi Wang},
  \bibinfo{person}{Dale Schuurmans}, \bibinfo{person}{Maarten Bosma},
  \bibinfo{person}{Brian Ichter}, \bibinfo{person}{Fei Xia},
  \bibinfo{person}{Ed Chi}, \bibinfo{person}{Quoc~V. Le}, {and}
  \bibinfo{person}{Denny Zhou}.} \bibinfo{year}{2022}\natexlab{b}.
\newblock \showarticletitle{Chain-of-{{Thought Prompting Elicits Reasoning}} in
  {{Large Language Models}}}.
\newblock \bibinfo{journal}{\emph{Advances in Neural Information Processing
  Systems}}  \bibinfo{volume}{35} (\bibinfo{date}{Dec.} \bibinfo{year}{2022}),
  \bibinfo{pages}{24824--24837}.
\newblock


\bibitem[Xiao et~al\mbox{.}(2025)]%
        {xiao_TradingAgentsMultiAgentsLLM_2025}
\bibfield{author}{\bibinfo{person}{Yijia Xiao}, \bibinfo{person}{Edward Sun},
  \bibinfo{person}{Di Luo}, {and} \bibinfo{person}{Wei Wang}.}
  \bibinfo{year}{2025}\natexlab{}.
\newblock \bibinfo{title}{{{TradingAgents}}: {{Multi-Agents LLM Financial
  Trading Framework}}}.
\newblock
\href{https://doi.org/10.48550/arXiv.2412.20138}{doi:\nolinkurl{10.48550/arXiv.2412.20138}}
\showeprint[arxiv]{2412.20138}~[q-fin]


\bibitem[Yang(2025)]%
        {yang_LLMsKnowWhen_2025}
\bibfield{author}{\bibinfo{person}{Lingyu Yang}.}
  \bibinfo{year}{2025}\natexlab{}.
\newblock \bibinfo{title}{Do {{LLMs Know When}} to {{Flip}} a {{Coin}}?
  {{Strategic Randomization}} through {{Reasoning}} and {{Experience}}}.
\newblock
\href{https://doi.org/10.48550/arXiv.2506.18928}{doi:\nolinkurl{10.48550/arXiv.2506.18928}}
\showeprint[arxiv]{2506.18928}~[cs]


\bibitem[Youngdahl et~al\mbox{.}(2024)]%
        {pactum2024}
\bibfield{author}{\bibinfo{person}{William Youngdahl}, \bibinfo{person}{Shail
  Patel}, {and} \bibinfo{person}{Addison Sutton}.}
  \bibinfo{year}{2024}\natexlab{}.
\newblock \bibinfo{booktitle}{\emph{Pactum's {{AI}} in {{Contract
  Negotiations}}: {{Walmart}} and {{Maersk}}}}.
\newblock \bibinfo{type}{{T}echnical {R}eport} TB0756-PDF-ENG.
  \bibinfo{institution}{Thunderbird School of Global Management, Arizona State
  University}.
\newblock


\bibitem[Zhang et~al\mbox{.}(2024)]%
        {zhang_KLevelReasoningEstablishing_2024}
\bibfield{author}{\bibinfo{person}{Yadong Zhang}, \bibinfo{person}{Shaoguang
  Mao}, \bibinfo{person}{Tao Ge}, \bibinfo{person}{Xun Wang},
  \bibinfo{person}{Yan Xia}, \bibinfo{person}{Man Lan}, {and}
  \bibinfo{person}{Furu Wei}.} \bibinfo{year}{2024}\natexlab{}.
\newblock \bibinfo{title}{K-{{Level Reasoning}}: {{Establishing Higher Order
  Beliefs}} in {{Large Language Models}} for {{Strategic Reasoning}}}.
\newblock
\href{https://doi.org/10.48550/arXiv.2402.01521v2}{doi:\nolinkurl{10.48550/arXiv.2402.01521v2}}
\showeprint[arxiv]{2402.01521v2}~[cs]


\bibitem[Zheng et~al\mbox{.}(2025)]%
        {zheng_NashEquilibriumBounded_2025}
\bibfield{author}{\bibinfo{person}{Kehan Zheng}, \bibinfo{person}{Jinfeng
  Zhou}, {and} \bibinfo{person}{Hongning Wang}.}
  \bibinfo{year}{2025}\natexlab{}.
\newblock \bibinfo{title}{Beyond {{Nash Equilibrium}}: {{Bounded Rationality}}
  of {{LLMs}} and Humans in {{Strategic Decision-making}}}.
\newblock
\href{https://doi.org/10.48550/arXiv.2506.09390}{doi:\nolinkurl{10.48550/arXiv.2506.09390}}
\showeprint[arxiv]{2506.09390}~[cs]


\end{thebibliography}
	
	\clearpage
	\onecolumn
	\appendix
	\section{Additional Experimental Results}
	
	\begin{table}[h]
		\centering
		\caption{LLMs used in this study and selected benchmark performances.}
		\label{tab:models}
		\begin{tabular}{l c c c}
			\toprule
			\textbf{Model} & \textbf{\#Params} & \textbf{GPQA-D (\%)} & \textbf{HLE (\%)} \\
			\midrule
			OpenAI o3                    & ---                          & 82.7  & 20.0 \\
			Claude Sonnet 4.5+           & 100B$^\dagger$         & 83.4  & 13.7 \\
			DeepSeek R1                  & 671B$^*$                     & 81.3  & 14.9 \\
			Qwen 3.5 397b+               & 397B$^*$                     & 89.3  & 27.3 \\
			\midrule[\heavyrulewidth]
			Claude Sonnet 4.5            & 100B$^\dagger$         & 83.4  & 7.5  \\
			DeepSeek V3                  & 685B$^*$                     & 65.5  & 5.2  \\
			Claude Haiku 4.5             & 10B$^\dagger$          & 73.0  & 4.3  \\
			Qwen 3.5-27b                 & 27B                          & 84.2  & 13.2 \\
			\midrule[\heavyrulewidth]
			Mercury 2                    & 7B$^\dagger$           & 77.0  & 15.5 \\
			\bottomrule
		\end{tabular}
		\parbox{\linewidth}{\footnotesize $^*$MoE architecture; active parameters per forward pass are 17B (Qwen 397b+), 37B (DeepSeek R1/V3). $^\dagger$Unofficial community estimate; not disclosed by provider. GPQA-D and HLE figures for the Claude family are from third-party leaderboards (Scale~Labs HLE leaderboard for HLE; Artificial Analysis / vals.ai for GPQA-D), since Anthropic does not publish these benchmarks for Claude Sonnet~4.5 or Claude Haiku~4.5 in their model cards.}
	\end{table}
	
	\begin{table}[h!]
		\centering
		\caption{Opponent identity and implied reasoning depth in BCG.}
		\begin{tabular}{lccccc}
			\toprule
			OPPONENT & (baseline) & Human & LLM & Expert & Yourself \\
			\midrule
			OpenAI o3 & \begin{tabular}{c}$L_2$ \\ (2.21 ± 0.14)\end{tabular} & \begin{tabular}{c}$L_2$ \\ (2.23 ± 0.14)\end{tabular} & \begin{tabular}{c}$L_\infty$ \\ (10.00 ± 0.29)\end{tabular} & \begin{tabular}{c}$L_\infty$ \\ (10.00 ± 0.29)\end{tabular} & \begin{tabular}{c}$L_\infty$ \\ (10.00 ± 0.29)\end{tabular} \\
			Claude Sonnet 4.5+ & \begin{tabular}{c}$L_3$ \\ (2.82 ± 0.05)\end{tabular} & \begin{tabular}{c}$L_2$ \\ (2.22 ± 0.04)\end{tabular} & \begin{tabular}{c}$L_4$ \\ (3.84 ± 0.06)\end{tabular} & \begin{tabular}{c}$L_5$ \\ (5.98 ± 0.24)\end{tabular} & \begin{tabular}{c}$L_\infty$ \\ (9.52 ± 0.16)\end{tabular} \\
			DeepSeek R1 & \begin{tabular}{c}$L_2$ \\ (2.50 ± 0.14)\end{tabular} & \begin{tabular}{c}$L_2$ \\ (2.18 ± 0.14)\end{tabular} & \begin{tabular}{c}$L_3$ \\ (4.52 ± 0.19)\end{tabular} & \begin{tabular}{c}$L_\infty$ \\ (9.67 ± 0.29)\end{tabular} & \begin{tabular}{c}$L_\infty$ \\ (10.00 ± 0.29)\end{tabular} \\
			Qwen 3.5 397b+ & \begin{tabular}{c}$L_2$ \\ (2.19 ± 0.14)\end{tabular} & \begin{tabular}{c}$L_2$ \\ (2.00 ± 0.13)\end{tabular} & \begin{tabular}{c}$L_3$ \\ (4.12 ± 0.19)\end{tabular} & \begin{tabular}{c}$L_\infty$ \\ (10.00 ± 0.29)\end{tabular} & \begin{tabular}{c}$L_\infty$ \\ (9.65 ± 0.29)\end{tabular} \\
			\midrule[\heavyrulewidth]
			Claude Sonnet 4.5 & \begin{tabular}{c}$L_3$ \\ (2.75 ± 0.05)\end{tabular} & \begin{tabular}{c}$L_3$ \\ (2.58 ± 0.05)\end{tabular} & \begin{tabular}{c}$L_3$ \\ (3.45 ± 0.05)\end{tabular} & \begin{tabular}{c}$L_4$ \\ (4.85 ± 0.19)\end{tabular} & \begin{tabular}{c}$L_\infty$ \\ (8.98 ± 0.24)\end{tabular} \\
			DeepSeek V3 & \begin{tabular}{c}$L_\infty$ \\ (9.02 ± 0.28)\end{tabular} & \begin{tabular}{c}$L_4$ \\ (5.44 ± 0.21)\end{tabular} & \begin{tabular}{c}$L_\infty$ \\ (10.00 ± 0.29)\end{tabular} & \begin{tabular}{c}$L_\infty$ \\ (10.00 ± 0.32)\end{tabular} & \begin{tabular}{c}$L_\infty$ \\ (10.00 ± 0.29)\end{tabular} \\
			Claude Haiku 4.5 & \begin{tabular}{c}$L_3$ \\ (4.22 ± 0.23)\end{tabular} & \begin{tabular}{c}$L_3$ \\ (3.09 ± 0.12)\end{tabular} & \begin{tabular}{c}$L_5$ \\ (5.49 ± 0.22)\end{tabular} & \begin{tabular}{c}$L_\infty$ \\ (8.95 ± 0.21)\end{tabular} & \begin{tabular}{c}$L_\infty$ \\ (7.62 ± 0.32)\end{tabular} \\
			Qwen 3.5-27b & \begin{tabular}{c}$L_2$ \\ (2.21 ± 0.14)\end{tabular} & \begin{tabular}{c}$L_2$ \\ (2.08 ± 0.13)\end{tabular} & \begin{tabular}{c}$L_4$ \\ (3.94 ± 0.18)\end{tabular} & \begin{tabular}{c}$L_\infty$ \\ (10.00 ± 0.29)\end{tabular} & \begin{tabular}{c}$L_\infty$ \\ (10.00 ± 0.29)\end{tabular} \\
			\midrule[\heavyrulewidth]
			Mercury 2 & \begin{tabular}{c}$L_\infty$ \\ (6.01 ± 0.22)\end{tabular} & \begin{tabular}{c}$L_2$ \\ (2.50 ± 0.14)\end{tabular} & \begin{tabular}{c}$L_\infty$ \\ (10.00 ± 0.29)\end{tabular} & \begin{tabular}{c}$L_\infty$ \\ (10.00 ± 0.29)\end{tabular} & \begin{tabular}{c}$L_\infty$ \\ (10.00 ± 0.29)\end{tabular} \\
			\bottomrule
		\end{tabular}
		\label{tab:BCG-bias}\\
		\tabnotes{Opponent identity and implied reasoning depth. Each cell shows the modal $L_k$; parentheses show the \emph{mean First-Terminal step} (capped at 10). $L_\infty$ indicates explicit jump to the limit-point (Nash) reasoning. Overthinking rates are reported separately in Table~\ref{tab:backtrack}. All opponent types are prompt-level identity framings: the tested model is told it plays against participants described as the indicated type; no separate agent participates in the experiment.}
	\end{table}

	\begin{table}[h]
		\centering
		\caption{BCG$^T$ backtracking behavior (n=100 trials).}
		\begin{tabular}{lccccc}
			\toprule
			Model & Mode Level & $\hat{\tau}$ (CH) & Total Steps & First Terminal & Overthinking \\
			\midrule
			OpenAI o3                & $L_3$ & $3.73 \pm 0.19$ & $4.0 \pm 0.2$ & $3.3 \pm 0.1$ & $24.0\%$ \\
			Claude Sonnet 4.5+       & $L_3$ & $3.31 \pm 0.18$ & $3.8 \pm 0.1$ & $3.6 \pm 0.1$ & $19.0\%$ \\
			DeepSeek R1              & $L_3$ & $3.53 \pm 0.19$ & $3.5 \pm 0.1$ & $3.4 \pm 0.1$ & $7.4\%$  \\
			Qwen 3.5 397b+           & $L_2$ & $2.03 \pm 0.14$ & $3.1 \pm 0.0$ & $2.4 \pm 0.1$ & $60.6\%$ \\
			\midrule[\heavyrulewidth]
			Claude Sonnet 4.5        & $L_3$ & $3.55 \pm 0.19$ & $4.4 \pm 0.1$ & $4.2 \pm 0.1$ & $16.0\%$ \\
			DeepSeek V3              & $L_4$ & $4.43 \pm 0.22$ & $4.3 \pm 0.1$ & $4.2 \pm 0.1$ & $0.0\%$  \\
			Claude Haiku 4.5         & $L_\infty$ & $5.32 \pm 0.23$ & $4.6 \pm 0.1$ & $4.1 \pm 0.2$ & $36.7\%$ \\
			Qwen 3.5-27b             & $L_2$ & $2.11 \pm 0.15$ & $3.5 \pm 0.1$ & $2.6 \pm 0.1$ & $42.4\%$ \\
			\midrule[\heavyrulewidth]
			Mercury 2                & $L_9$ & $10.00 \pm 0.32$ & $8.0 \pm 0.4$ & $6.2 \pm 0.2$ & $2.0\%$  \\
			\bottomrule
		\end{tabular}
		\label{tab:backtrack-full}\\
		\tabnotes{
			Reported values are sample means $\pm$ standard errors (across $n=100$ trials), 
			except for $\hat{\tau}$, where the asymptotic standard error of the Poisson MLE is 
			$\scriptstyle\sqrt{\hat{\tau}/n}$. 
			\emph{First Terminal} is the first step index at which the final action appears in the reasoning trace. 
			\emph{Overthinking} is the proportion of trials where the model revisited earlier levels after reaching its final decision: $\scriptstyle\frac{1}{n}\sum_{t=1}^n \indicator [\mathrm{FT}^{(t)} < \mathrm{Total}^{(t)}]\times 100$.
		}
	\end{table}

	\begin{table}[h]
		\centering
		\caption{Distributional distance to reference distributions.}
		
		\begin{tabular}{lcccccccc}
			\toprule
			& \multicolumn{4}{c}{\textbf{Expert vs Nash}} & \multicolumn{4}{c}{\textbf{Human vs Paper}} \\
			Model & KL & TV & $\ell_2$ & EMD & KL & TV & $\ell_2$ & EMD \\
			\midrule
			OpenAI o3 & 2.37 & 0.65 & 0.64 & 1.65 & 1.00 & 0.55 & 0.63 & 1.24 \\
			Claude Sonnet 4.5+ & 2.96 & 0.57 & 0.42 & 1.37 & 0.55 & 0.45 & 0.43 & 0.78 \\
			DeepSeek R1 & 1.44 & 0.43 & 0.41 & 1.08 & 1.68 & 0.79 & 0.89 & 1.75 \\
			Qwen 3.5 397b+ & 3.89 & 0.54 & 0.44 & 1.66 & 1.18 & 0.65 & 0.73 & 1.55 \\
			\midrule[\heavyrulewidth]
			Claude Sonnet 4.5 & 5.49 & 0.51 & 0.47 & 2.58 & 0.62 & 0.42 & 0.33 & 0.74 \\
			DeepSeek V3 & 26.91 & 0.98 & 1.05 & 5.54 & 4.53 & 0.75 & 0.77 & 4.64 \\
			Claude Haiku 4.5 & 12.41 & 0.92 & 0.97 & 5.76 & 1.94 & 0.72 & 0.75 & 4.39 \\
			Qwen 3.5-27b & 3.11 & 0.80 & 0.85 & 2.58 & 1.53 & 0.74 & 0.81 & 1.55 \\
			\midrule[\heavyrulewidth]
			Mercury 2 & 4.55 & 0.65 & 0.56 & 2.05 & 1.60 & 0.77 & 0.68 & 1.83 \\
			\bottomrule
		\end{tabular}
		
		\tabnotes{KL = Kullback–Leibler divergence; TV = total variation distance; $\ell_2$ = Euclidean distance; EMD = Earth Mover’s Distance.}
	\end{table}

	\begin{figure}[h]
		\centering
		\caption{Claude Sonnet 4.5+ BCG experimental results. }
		\includegraphics[width=\linewidth]{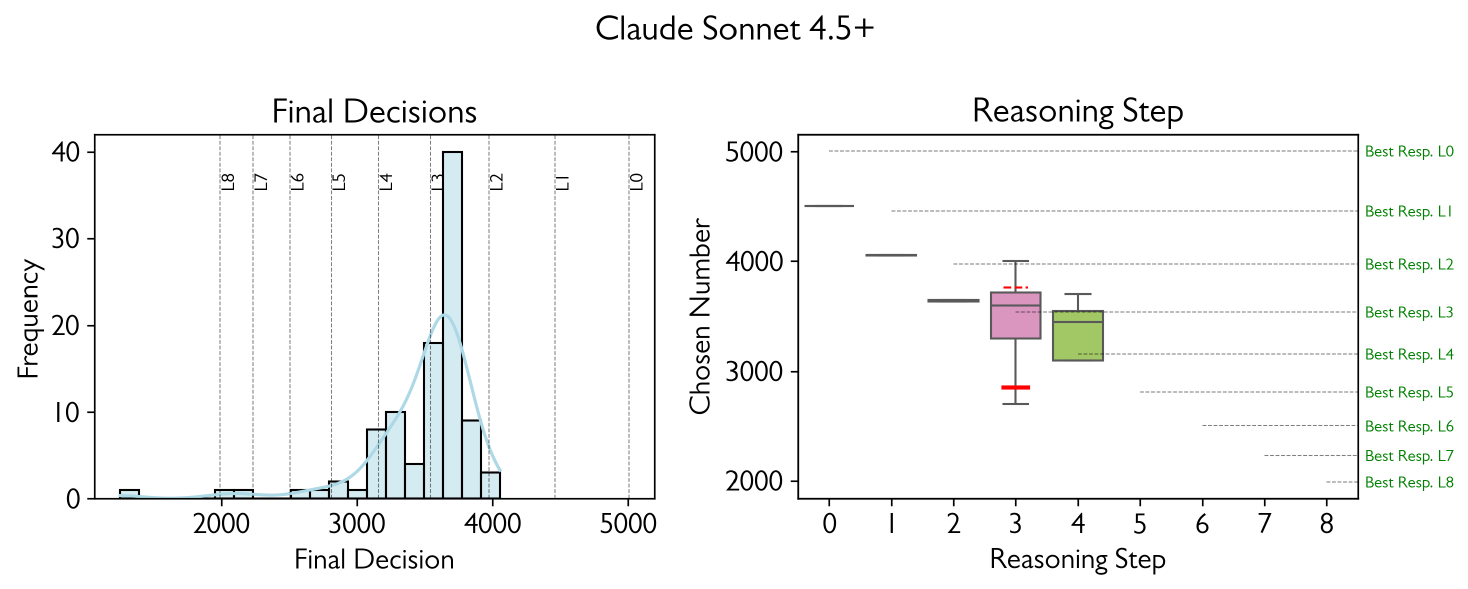}
		
		\label{fig:bcg-claude}
		\tabnotes{(left) Distribution of final chosen numbers across trials with implied level‑k reference lines. (right) Distribution of chosen numbers at each reasoning step. Red lines show the 95\% confidence interval which are flat in this case as the model consistently picks the right number. Annotated percentages indicate the proportion of trials that reached at least that step, reflecting how deeply the model typically chooses to reason.}
	\end{figure}
	
	\begin{figure}
		\centering
		\caption{BCG free-play final decisions mapped to level-$k$ buckets.}

		\includegraphics[width=\linewidth]{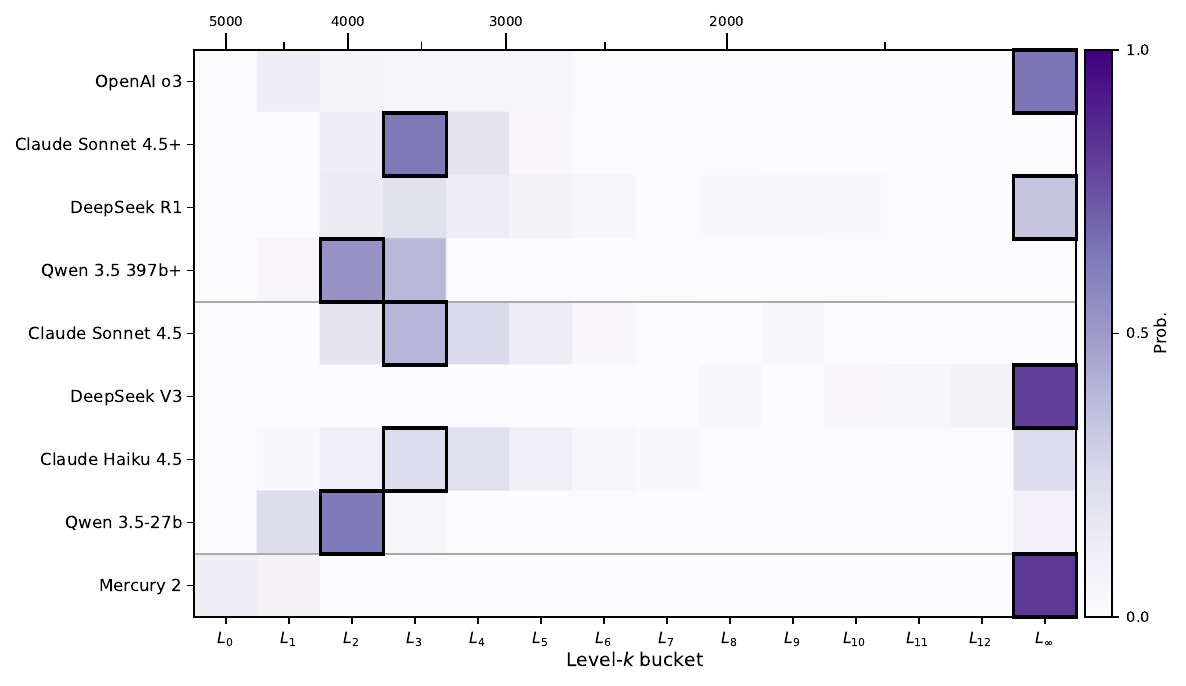}

		\label{fig:BCG1-appendix}
		\tabnotes{Each cell shows the fraction of trials (out of 100) whose final decision was closest to the corresponding level-$k$ target value. Target values (shown on the top axis) are computed as the fixed-point best response to level-$(k-1)$ play; $L_\infty$ captures responses at or near the lower bound (1250). Black rectangle marks the modal bucket for each model. Grey lines separate reasoning models (top), standard models (middle), and diffusion models (bottom).}
	\end{figure}
	
	\begin{figure}
		\centering
		\caption{Action distributions in the MRG: expert framing (vs.\ Nash equilibrium) and human framing (vs.\ human play).}

		\includegraphics[width=\linewidth]{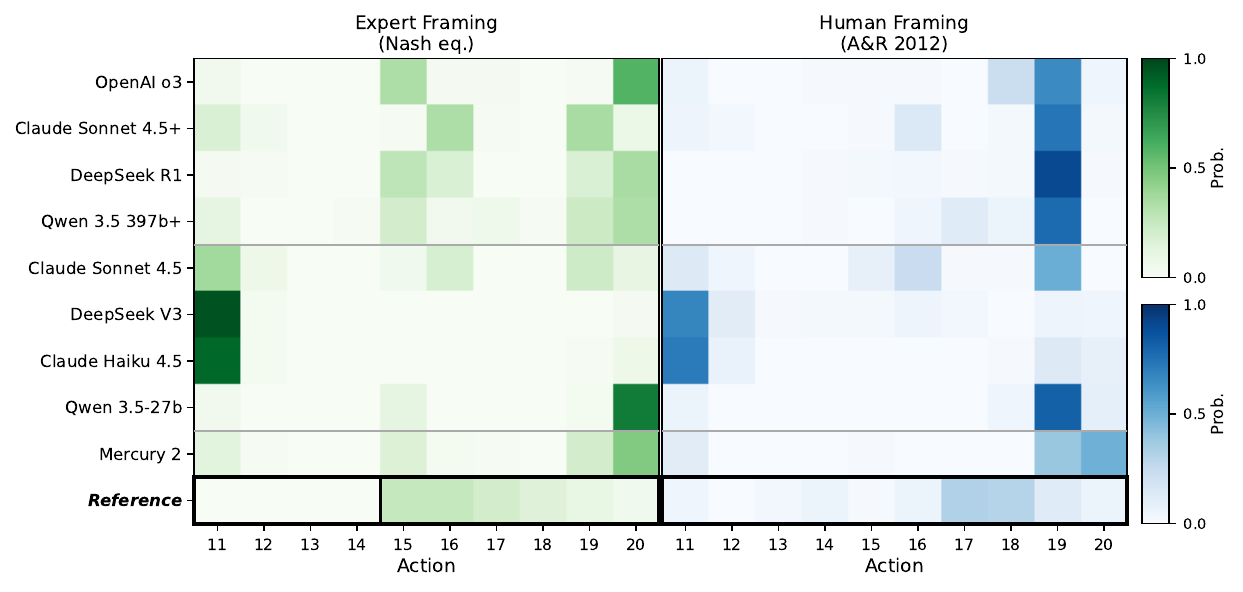}

		\label{fig:MRG-comparison}
		\tabnotes{Each cell shows the probability that a model (row) chose a given action (column). \textit{Reference} row (boxed) shows the benchmark distribution. Left panel: Nash mixed-strategy equilibrium with support $S^+=\{15,\ldots,20\}$ (highlighted). Right panel: empirical human play from Arad~\&~Rubinstein (2012).}
	\end{figure}

	\begin{figure}
		\centering
		\caption{Calculation error at various relative tolerance levels $\tol$: many of the smaller models make calculation errors which end up not giving the best-response.}
		
		\includegraphics[width=1\textwidth]{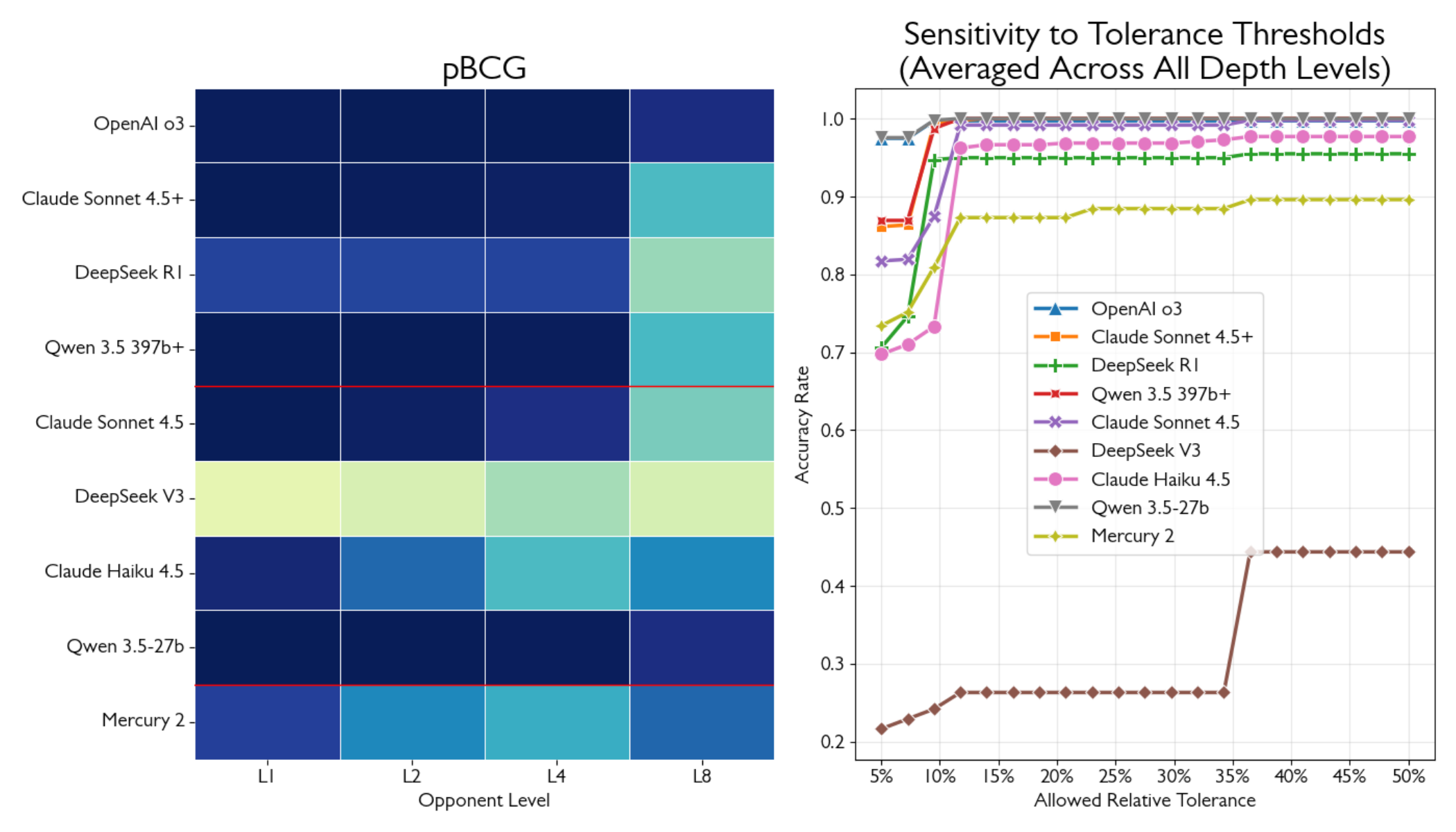}
		\label{fig:calc-error}
	\end{figure}

	\begin{figure}
		\centering
		\caption{\textbf{UMG: theoretical level-$k$ chain and empirical equilibrium selection.} Most models concentrate on the $(A,\,M)$ equilibrium (green), with a minority selecting $(F,\,P)$ (blue).}
		\includegraphics[width=\textwidth]{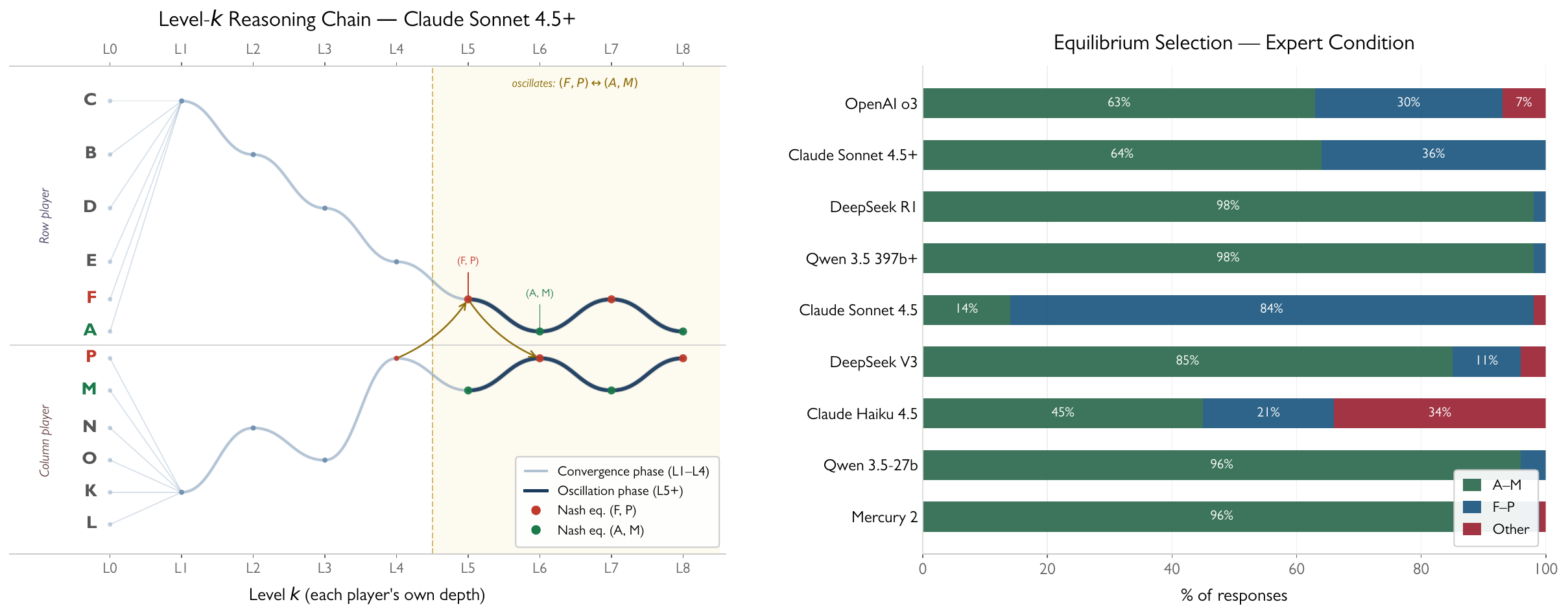}
		\label{fig:UMG-expert}
		\tabnotes{
			(\textit{Left}) Level-$k$ reasoning chain for the UMG (Claude Sonnet~4.5+).
			Each player is shown from their \emph{own} perspective: the x-axis is level $k$ for both players independently; the top band shows Row player's strategy at depth $k$ and the bottom band shows Column player's strategy at depth $k$.
			Standard level-$k$: Row at $L_k$ best-responds to Column at $L_{k-1}$ (one tick to the left), and vice versa.
			Both bands begin from $L_0 =$ uniform (grey bar).
			During the convergence phase ($L_1$--$L_4$, faint), each player's strategy descends toward a Nash equilibrium.
			From $L_5$ onward, Row oscillates $F \leftrightarrow A$ and Column oscillates $M \leftrightarrow P$ (both every two levels), never settling on the same Nash equilibrium simultaneously.
			Gold arrows trace one mutual-response cycle: Column's $L_4 = P$ leads Row to play $F$ at $L_5$ (payoff 94, Nash $(F,P)$); Row's $L_5 = F$ leads Column to play $P$ again at $L_6$ --- sustaining the oscillation.	\footnotemark

			(\textit{Right}) Empirical equilibrium selection under the expert condition across all models in the panel.
			}
	\end{figure}
    \footnotetext[\value{footnote}]{At $L_5$, Row reasons: ``Column at $L_4$ plays $P$; my best response to $P$ is $F$ (payoff 94).'' At $L_6$, Column reasons: ``Row at $L_5$ plays $F$; my best response to $F$ is $P$ (payoff 94).'' This mutual best-response cycle means neither player's sequence converges: Row's $F$ at odd levels $\geq 5$ and $A$ at even levels $\geq 6$ alternate indefinitely, as does Column's $M$/$P$.}
    
	\begin{figure}
		\centering
		\caption{\textbf{Human vs Expert Vocabulary}: Vocabulary shifts with opponent sophistication. Each row shows the usage rate difference (vs.\ expert) $-$ (vs.\ human) for a keyword group, computed from Claude Sonnet 4.5+ traces in the MRG Bias condition (100 trials per opponent type). Positive values (right) indicate vocabulary more common when facing a game theory expert; negative values (left) indicate vocabulary more common when facing a human. Rates are shown as (expert\% $|$ human\%) per keyword, with the dominant condition in black. All differences are statistically significant (95\% bootstrap CI, $p < 0.05$).}
		
		\includegraphics[width=\linewidth]{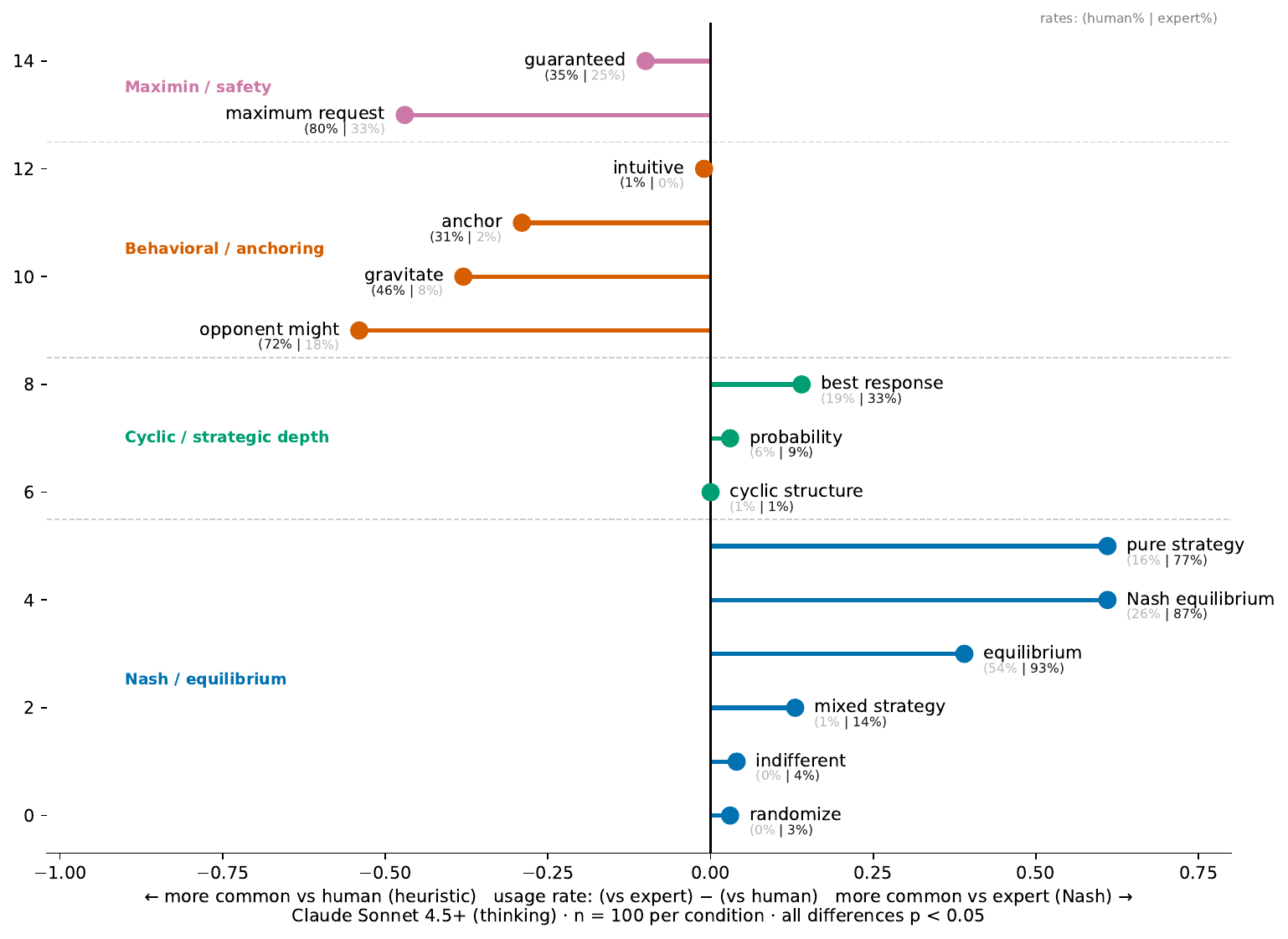}

		\label{fig:heuristic-text}
	\end{figure}
	
	\begin{table}[t]
		\centering
		\caption{Cross-game examples of reasoning heuristics in Claude Sonnet 4.5+ traces, organized by belief formation, evaluation, and choice. Examples are quoted as model rationales, not as validated equilibrium analyses. Cell provenance is given as scenario / opponent framing / trial id.}
		\label{tab:heuristics_crossgame}
		\footnotesize
		\begin{tabularx}{\textwidth}{p{2.9cm} p{3.0cm} X}
			\toprule
			Heuristic & Mechanism & Representative evidence \\
			\midrule
			
			\multicolumn{3}{l}{\textit{Belief formation: how the model forms conjectures about opponents}} \\
			\midrule
			
			Opponent-type depth anchoring
			& Opponent labels fix expected sophistication.
			&
			\textit{BCG-Bias / expert / 67:}
			``Game theory experts in one-shot games like this typically reason to level 4--6 or recognize the Nash equilibrium.'' \newline
			\textit{MRG-Bias / human / 13:}
			``Humans often employ limited levels of strategic thinking: Level 0 thinker: might choose 20 (maximizing direct payment); Level 1 thinker: realizes they should undercut 20-choosers, so picks 19; Level 2 thinker: anticipates level 1 thinking, so picks 18.'' \newline
			\textit{UMG / human / 60:}
			``Since humans typically reason at Level-1 or Level-2 in one-shot games \ldots\ I'll assume Player 2 is likely Level-2, which means I should play Level-3.'' \\
			
			Behavioral opponent modeling
			& Models opponents as a mixture or menu of likely behavioral types.
			&
			\textit{BCG-Bias / human / 6:}
			``I expect the population distribution to be roughly: \(\sim\)20\% will choose high numbers (40--50) with naive/level-0 thinking; \(\sim\)30\% will choose around 30--35 with level-1 thinking; \(\sim\)30\% will choose around 20--25 with level-2 thinking; \(\sim\)20\% will choose around 15 or lower with level-3+ thinking.'' \newline
			\textit{MRG-Bias / human / 80:}
			``Some players might anchor on the maximum (20) as a \emph{safe} choice; others might choose 19, anticipating opponents will choose 20; the game could spiral into lower numbers if both players think through multiple levels.'' \newline
			\textit{UMG / human / 1:}
			``If Player 2 is Level-1 \ldots\ they would choose K. Against K, my best response is B (payoff 77). \ldots\ If Player 2 is Level-2, they expect me to choose C \ldots\ they would choose N. Against N, my best response is D (payoff 82). \ldots\ If Player 2 is Level-3, they expect me to choose B \ldots\ they would choose O. Against O, I would choose E (payoff 85).'' \\
			
			\midrule
			\multicolumn{3}{l}{\textit{Evaluation: how the model maps conjectures into candidate actions}} \\
			\midrule
			
			Local best-response step
			& Infers a likely opponent action, then computes the local best response.
			&
			\textit{BCG-Bias / human / 27:}
			``If others average $M$, my optimal choice is: $x = (20/31) \times M$. If others average 27, I should pick: $(20/31) \times 27 \approx 17.4$.'' \newline
			\textit{MRG-Bias / human / 60:}
			``For any amount my opponent chooses (call it $X$), my best response would be to choose $X-1$, gaining $X-1+20 = X+19$ shekels.'' \newline
			\textit{UMG / expert / 19:}
			``If Player 2 is Level-2, they expect me to play B (from Level-1) \ldots\ Player 2's best response is N (payoff 81). Against N, my best response is D (payoff 82).'' \\
			
			Equilibrium / fixed-point analysis
			& Formal solution concepts, cycles, or mutual best-response fixed points are used to structure evaluation.
			&
			\textit{BCG-Bias / expert / 39:}
			``This game has a unique Nash equilibrium at 0. \ldots\ If everyone picks 0, the mean is 0, and $2/3 \times 0 = 0$. So 0 is the best response to everyone else picking 0.'' \newline
			\textit{MRG-Bias / expert / 20:}
			``Nash Equilibrium Analysis: \ldots\ If opponent chooses $X$ (where $X \geq 12$), my best response is $X-1$ \ldots\ But if I choose $X-1$, their best response is $X-2$. This creates a cycle with no stable equilibrium.'' \newline
			\textit{UMG / expert / 32:}
			``(A,~M) is a Nash equilibrium: given Player 2 plays M, A gives me 96 (my best response); given I play A, M gives Player 2 96 (their best response).'' \\
			
			\midrule
			\multicolumn{3}{l}{\textit{Choice: how the model selects when recursion is unstable or indeterminate}} \\
			\midrule
			
			Finite-depth fallback 
			& Recognizes formal, cyclic, or recursive logic, then commits to a finite-depth, empirical, or focal action.
			&
			\textit{BCG free-play / no framing / 24:}
			``This creates an iterative process: $5005 \to 4505 \to 4055 \to 3650 \to 3285$, etc. The Nash equilibrium would be at the minimum (1250), but in practice, people rarely reason that far.'' \newline
			\textit{MRG-Bias / human / 98:}
			``This creates a `race to the bottom' dynamic that would theoretically push both players toward the minimum (11). However, in practice with human players: humans rarely reason all the way to such extremes.'' \newline
			\textit{UMG / opponent / 53:}
			``Most sophisticated players operate around Level-2 to Level-4 in practice. \ldots\ playing E anticipates that Player 2 (at Level-3) will play O, yielding me a strong payoff of 85.'' \\
			
			Focal / payoff-salient selection
			& Uses salience or high joint payoff to resolve indeterminacy.
			&
			\textit{BCG free-play / no framing / 85:}
			``I'm choosing 3400 as it targets the sweet spot where players have done 2--3 levels of iterative reasoning from the midpoint. This assumes a moderately sophisticated player pool without going to the game-theoretic equilibrium.'' \newline
			\textit{MRG-Bias / expert / 17:}
			``The natural focal point is the \emph{maximum value of 20}, because: 1. it's a clear boundary/focal point; 2. it guarantees a payoff of 20 with no risk; 3. if we both try to \emph{outsmart} each other by going lower, we both lose.'' \newline
			\textit{UMG / expert / 20:}
			``The (A,~M) equilibrium is focal because: 1. it's Pareto optimal (both get 96, the maximum achievable in any cell); 2. it's the most symmetric and salient outcome; 3. high-level reasoning converges toward it.'' \\
			
			Robustness under non-convergence
			& Chooses an action that performs well across nearby predicted opponent moves or unstable recursive paths.
			&
			\textit{BCG free-play / no framing / 46:}
			``Given uncertainty in my estimate, I'll choose 3400 as it's robust across a reasonable range of assumptions (if others average 3500--4000, optimal choices range from 3100--3600).'' \newline
			\textit{MRG-Bias / expert / 31:}
			``A game theory expert would think one level ahead of `choosing maximum' and would choose 19 to exploit someone choosing 20. But knowing this, I should also choose 19, as it's the robust choice that performs well against the most likely opponent strategies.'' \newline
			\textit{UMG / human / 3:}
			``Since humans typically exhibit Level-1 to Level-3 reasoning in these games, and Level-3 gives me a strong payoff of 82 (and D also performs well against P with 92), I'll go with the Level-3 strategy.'' \\
			
			\bottomrule
		\end{tabularx}
		
		\tabnotes{All quotations are verbatim from Claude Sonnet 4.5+ (extended thinking) reasoning traces at $T=0.75$. Scenario keys: BCG-Bias = \texttt{pBCG\_bias}; MRG-Bias = \texttt{pMRG\_bias}; UMG = \texttt{ladder-expert-eq}; BCG free-play = \texttt{pBCG\_paper\_w\_tracing}. Trial ids index the corresponding \texttt{final.json}.}
	\end{table}
	
	\begin{table}[h]
		\centering
		\caption{MRG keyword usage by Nash-support membership across the current model panel, split by opponent framing.}
		\label{tab:keywords}
		\footnotesize
		\begin{tabular}{llcccccccc}
			\toprule
			& & \multicolumn{2}{c}{Nash / equilibrium} & \multicolumn{2}{c}{Cyclic / depth} & \multicolumn{2}{c}{Behavioral / anchoring} & \multicolumn{2}{c}{Maximin / safety} \\
			\cmidrule(lr){3-4}\cmidrule(lr){5-6}\cmidrule(lr){7-8}\cmidrule(lr){9-10}
			Model & Opp.\ framing & $S^+$ & $S^-$ & $S^+$ & $S^-$ & $S^+$ & $S^-$ & $S^+$ & $S^-$ \\
			\midrule
			\multirow{2}{*}{OpenAI o3} & Expert & 0 (0\%) & 0 (0\%) & 0 (0\%) & 0 (0\%) & 0 (0\%) & 0 (0\%) & 0 (0\%) & 0 (0\%) \\
			& Human & 0 (0\%) & 0 (0\%) & 0 (0\%) & 0 (0\%) & 0 (0\%) & 0 (0\%) & 0 (0\%) & 0 (0\%) \\
			\addlinespace[1pt]
			\multirow{2}{*}{Claude Sonnet 4.5+} & Expert & 77 (77\%) & 16 (16\%) & 29 (29\%) & 9 (9\%) & 20 (20\%) & 3 (3\%) & 23 (23\%) & 3 (3\%) \\
			& Human & 49 (49\%) & 5 (5\%) & 21 (21\%) & 4 (4\%) & 88 (88\%) & 5 (5\%) & 37 (37\%) & 4 (4\%) \\
			\addlinespace[1pt]
			\multirow{2}{*}{DeepSeek R1} & Expert & 92 (92\%) & 3 (3\%) & 75 (75\%) & 1 (1\%) & 36 (36\%) & 2 (2\%) & 45 (45\%) & 2 (2\%) \\
			& Human & 38 (38\%) & 1 (1\%) & 50 (50\%) & 1 (1\%) & 95 (95\%) & 1 (1\%) & 72 (72\%) & 1 (1\%) \\
			\addlinespace[1pt]
			\multirow{2}{*}{Qwen 3.5 397b+} & Expert & 40 (40\%) & 9 (9\%) & 36 (36\%) & 7 (7\%) & 14 (14\%) & 2 (2\%) & 25 (25\%) & 2 (2\%) \\
			& Human & 26 (26\%) & 1 (1\%) & 15 (15\%) & 0 (0\%) & 41 (41\%) & 1 (1\%) & 7 (7\%) & 0 (0\%) \\
			\midrule[\heavyrulewidth]
			\multirow{2}{*}{Claude Sonnet 4.5} & Expert & 56 (56\%) & 44 (44\%) & 33 (33\%) & 36 (36\%) & 36 (36\%) & 16 (16\%) & 22 (22\%) & 11 (11\%) \\
			& Human & 83 (83\%) & 17 (17\%) & 41 (41\%) & 7 (7\%) & 82 (82\%) & 15 (15\%) & 29 (29\%) & 3 (3\%) \\
			\addlinespace[1pt]
			\multirow{2}{*}{DeepSeek V3} & Expert & 2 (2\%) & 98 (98\%) & 2 (2\%) & 64 (64\%) & 0 (0\%) & 21 (21\%) & 0 (0\%) & 10 (10\%) \\
			& Human & 17 (17\%) & 80 (80\%) & 7 (7\%) & 23 (23\%) & 19 (19\%) & 75 (75\%) & 5 (5\%) & 21 (21\%) \\
			\addlinespace[1pt]
			\multirow{2}{*}{Claude Haiku 4.5} & Expert & 8 (8\%) & 92 (92\%) & 4 (4\%) & 44 (44\%) & 2 (2\%) & 12 (12\%) & 4 (4\%) & 10 (10\%) \\
			& Human & 19 (19\%) & 73 (73\%) & 8 (8\%) & 14 (14\%) & 21 (21\%) & 53 (53\%) & 16 (16\%) & 24 (24\%) \\
			\addlinespace[1pt]
			\multirow{2}{*}{Qwen 3.5-27b} & Expert & 76 (76\%) & 4 (4\%) & 94 (94\%) & 4 (4\%) & 13 (13\%) & 2 (2\%) & 78 (78\%) & 0 (0\%) \\
			& Human & 30 (30\%) & 6 (6\%) & 88 (88\%) & 3 (3\%) & 84 (84\%) & 3 (3\%) & 62 (62\%) & 4 (4\%) \\
			\midrule[\heavyrulewidth]
			\multirow{2}{*}{Mercury 2} & Expert & 0 (0\%) & 0 (0\%) & 0 (0\%) & 0 (0\%) & 0 (0\%) & 0 (0\%) & 0 (0\%) & 0 (0\%) \\
			& Human & 0 (0\%) & 0 (0\%) & 0 (0\%) & 0 (0\%) & 0 (0\%) & 0 (0\%) & 0 (0\%) & 0 (0\%) \\
			\bottomrule
		\end{tabular}\\
		\tabnotes{Keyword occurrence in reasoning traces in the Money Request Game (MRG) bias scenario (\texttt{pMRG\_bias}), with the model told it is playing against either a game-theory expert or a human ($n=100$ trials per model and condition). For each model, opponent framing, and keyword group, we report the number of traces (and \% of the trace pool) where at least one term in the group is mentioned, partitioned by whether the chosen action falls within the Nash equilibrium support $S^+ = \{15,\ldots,20\}$ or strictly outside it, $S^- = \{11,\ldots,14\}$. Group definitions follow Fig.~\ref{fig:heuristic-text}: \emph{Nash / equilibrium} (equilibrium, Nash, mixed-/pure-strategy, randomize, indifferent); \emph{Cyclic / strategic depth} (best-response, probability, cyclic); \emph{Behavioral / anchoring} (anchor, gravitate, intuitive, opponent/human/player + might/likely/tend); \emph{Maximin / safety} (`maximum' only when co-mentioning safe/guarantee/risk/hedge within $\sim$30 characters; and bare `guaranteed'). Being within $S^+$ does not imply playing a unique best response to any specific opponent action. OpenAI~o3 and Mercury~2 do not emit verbalized reasoning text in this scenario (their responses are limited to the final \texttt{<response>}~tag); their rows therefore record zero matches across all groups.}
	\end{table}
	
	\begin{figure}
		\centering
		\caption{Emergence of heuristics in larger models}    
		\includegraphics[width=\textwidth]{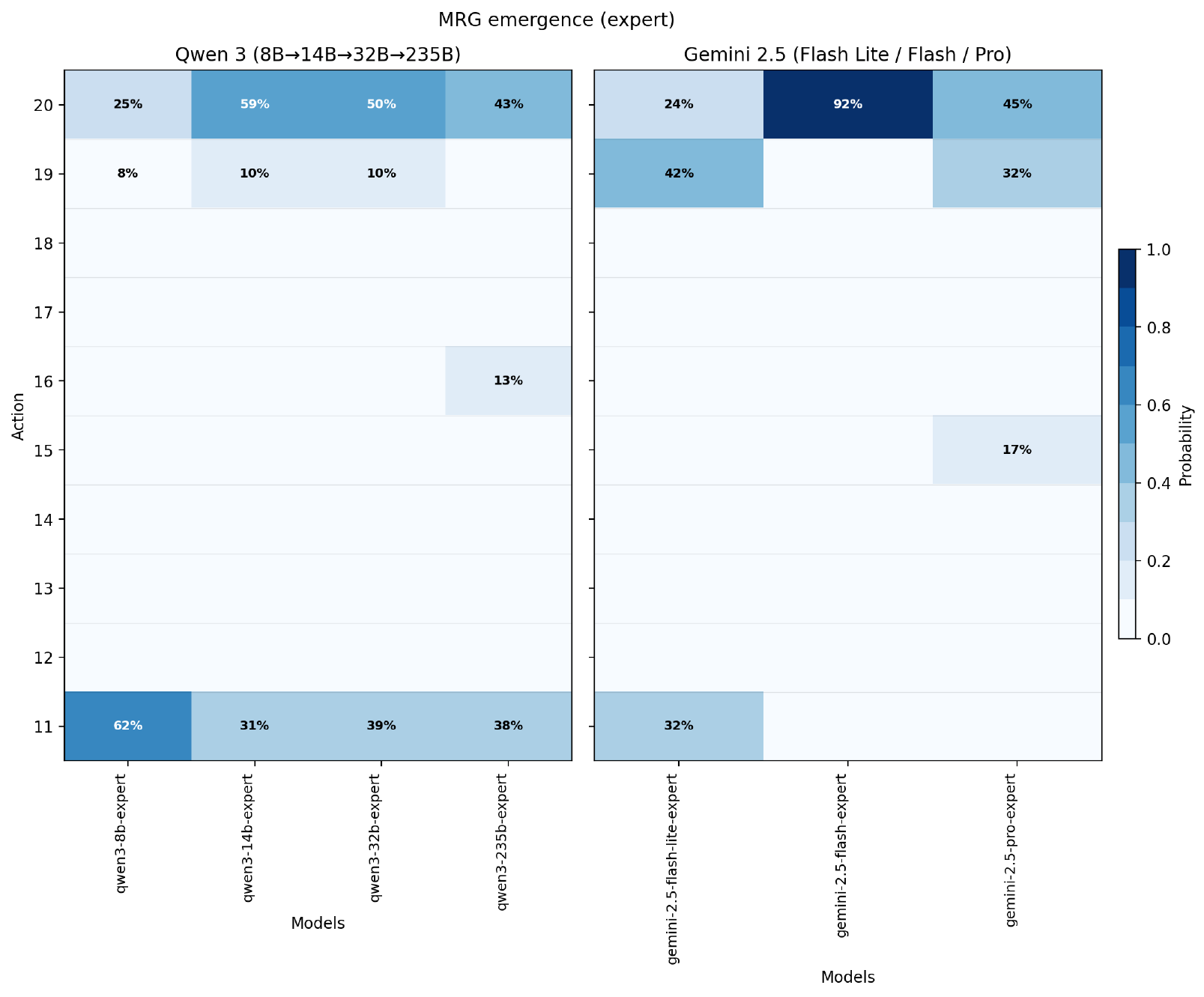}
		\label{fig:MRG-emergence-expert}
	\end{figure}
	\clearpage

	% ============================================================
% Symbol macros — requires \usepackage{amssymb} (checkmark) and
% \usepackage{nicefrac} (half); both are standard in NeurIPS.
\newcommand{\cmark}{\checkmark}
\newcommand{\xmark}{$\times$}
\newcommand{\hmark}{$\cdot$}

\clearpage
\section{Prior Knowledge Check}
\label{sec:appendix-knowledge}

\setcounter{table}{0}
\renewcommand{\thetable}{S\arabic{table}}
\setcounter{figure}{0}
\renewcommand{\thefigure}{S\arabic{figure}}

Each model was presented with the three experimental game descriptions from the experiments. We then asked four questions from prior knowledge only to check the prior knowledge of the LLMs. Table~\ref{tab:knowledge-key} lists the correct answers; Table~\ref{tab:knowledge} reports model scores (\cmark~=~correct, \xmark~=~incorrect, \hmark~=~partial).

\begin{table}[h]
	\centering
	\caption{Answer key for the prior knowledge check. A checkmark in Table~\ref{tab:knowledge} means the model gave the answer listed here. For UMG, this means correctly reporting no prior knowledge.}
	\label{tab:knowledge-key}
	\small
	
	\begin{tabular}{l p{2.4cm} p{2.6cm} p{3.4cm} p{2.6cm} p{3.0cm}}
		\toprule
		Game & Q0: Named recognition & Q1: Description recognition & Q2: Nash equilibrium & Q3: Human behavior & Q4: Reference \\
		\midrule
		BCG
		& Recognizes as \textit{p}-beauty contest
		& Recognizes as \textit{p}-beauty contest
		& Lower bound (1250)
		& $\approx$20--35 (L1--L2)
		& \cite{nagel_UnravelingGuessingGames_1995} \\[2pt]
		
		MRG
		& Recognizes as 11--20 money request game
		& Recognizes as 11--20 money request game
		& No PSNE; mixed support $\{15,\ldots,20\}$
		& Most common = 20
		& \cite{arad_1120MoneyRequest_2012} \\[2pt]
		
		UMG
		& Correctly reports no known named game
		& Correctly reports no known payoff matrix
		& Correctly reports no known NE from prior literature
		& Correctly reports no published human data
		& Correctly reports no published reference \\
		\bottomrule
	\end{tabular}
\end{table}

\begin{table}[h]
	\centering
	%\caption{Prior knowledge check scores (\cmark~matches the answer key, \xmark~contradicts the answer key, \hmark~partial or unstable, --~no response). Q0 is named-game recognition before seeing any description.}
    \caption{Prior knowledge check scores. \cmark~matches the answer key, \xmark~contradicts it, \hmark~is partial or unstable, and -- indicates no response. Q0 tests named-game recognition before any description.}
	\label{tab:knowledge}
	\small
	
	\begin{tabular}{l ccccc ccccc ccccc}
		\toprule
		& \multicolumn{5}{c}{\textbf{BCG}} & \multicolumn{5}{c}{\textbf{MRG}} & \multicolumn{5}{c}{\textbf{UMG}} \\
		\cmidrule(lr){2-6}\cmidrule(lr){7-11}\cmidrule(lr){12-16}
		Model & Q0 & Q1 & Q2 & Q3 & Q4 & Q0 & Q1 & Q2 & Q3 & Q4 & Q0 & Q1 & Q2 & Q3 & Q4 \\
		\midrule
		OpenAI o3
		& \cmark & \cmark & \cmark & \cmark & \cmark
		& \xmark & \xmark & \xmark & \cmark & \xmark
		& \hmark & \cmark & \cmark & \cmark & \cmark \\
		Claude Sonnet 4.5+% TODO: re-run knowledge check on new claude family if budget permits — values inherited from 3.7 generation
		& \cmark & \cmark & \cmark & \cmark & \cmark
		& \xmark & \xmark & \xmark & \xmark & \xmark
		& \cmark & \cmark & \cmark & \cmark & \cmark \\
		DeepSeek R1
		& \cmark & \cmark & \cmark & \cmark & \hmark
		& \hmark & \xmark & \xmark & \cmark & \xmark
		& \hmark & \cmark & \cmark & \cmark & \cmark \\
		Qwen 3.5 397b+
		& \cmark & \cmark & \cmark & \cmark & \cmark
		& \hmark & \cmark & \hmark & \xmark & \cmark
		& \cmark & \cmark & \cmark & \cmark & \cmark \\
		\midrule[\heavyrulewidth]
		Claude Sonnet 4.5
		& \cmark & \cmark & \cmark & \cmark & \cmark
		& \xmark & \xmark & \xmark & \xmark & \xmark
		& \cmark & \cmark & \cmark & \cmark & \cmark \\
		DeepSeek V3
		& \cmark & \cmark & \cmark & \cmark & \cmark
		& \xmark & \xmark & \xmark & \hmark & \xmark
		& \hmark & \cmark & \cmark & \cmark & \cmark \\
		Claude Haiku 4.5
		& \cmark & \cmark & \xmark & \cmark & \cmark
		& \xmark & \xmark & \xmark & \xmark & \xmark
		& \cmark & \cmark & \cmark & \cmark & \cmark \\
		Qwen 3.5-27b
		& \cmark & \cmark & \cmark & \cmark & --
		& \xmark & \xmark & \hmark & \hmark & \xmark
		& \cmark & \cmark & \cmark & \cmark & \cmark \\
		\midrule[\heavyrulewidth]
		Mercury 2
		& \cmark & \cmark & \cmark & \hmark & \hmark
		& \xmark & \cmark & \xmark & \cmark & \xmark
		& \hmark & \cmark & \cmark & \cmark & \cmark \\
		\bottomrule
	\end{tabular}\\
	\tabnotes{
        Q0--Q4 test, respectively: named-game recognition, recognition from anonymized description, Nash equilibrium from memory, typical human behavior, and key academic reference. For UMG, \cmark~means correct non-recognition from prior literature. Partial marks indicate unstable, generic, or materially incomplete answers.
    }
\end{table}

\noindent \textbf{Conclusion}:
\begin{enumerate}
	\item \textit{BCG is universally recognized.} Every model correctly identifies the game; nearly all recall the NE at the lower bound and cite~\citet{nagel_UnravelingGuessingGames_1995}, confirming a memorization risk.
	
	\item \textit{MRG is partially known.} Two models (Mercury~2 and Qwen~3.5~397b+) correctly identify the game by name; only Qwen~3.5~397b+ also cites the correct reference \citep{arad_1120MoneyRequest_2012}. No model recalls the correct NE support $\{15,\ldots,20\}$. Models that misidentify the game as the Traveler's Dilemma nevertheless correctly report modal human choice (20), since this pattern characterizes both games.
	
	\item \textit{UMG is universally unrecognized.} No model identifies the payoff matrix or names a paper, consistent with the game having been introduced in this paper.
\end{enumerate}
	\section{Experiment Prompts}
\label{sec:appendix-prompt}

Each experiment is delivered as a two-turn exchange: a \textbf{system} prompt (role/instructions) followed by a \textbf{user} prompt (game description).
Template placeholders are filled per experiment before the call is issued; the \texttt{emph} entries such as \texttt{\{COMPETITION\_DEPTH\}} show the variable names as sent.
No level-$k$ language appears in the tracing scenario (\S\ref{sec:appendix-prompt-bcg-tracing}), which measures spontaneous reasoning depth.

\subsection{Beauty Contest Game (BCG)}

\subsubsection*{Level-$k$ depth (Figure~\ref{fig:model-capability})}

Scenario \texttt{pBCG\_depth}.
$p = 0.9$, $n = 11$, range $= [1250, 8761]$.
Conditions: \texttt{COMPETITION\_DEPTH} $\in \{\mathrm{L1}, \mathrm{L2}, \mathrm{L4}, \mathrm{L8}\}$.

\begin{promptbox}[pBCG Depth Prompt]
(*\systag*)

You are a participant in a one-shot economic decision task. You can assume that the other participants are level {COMPETITION_DEPTH} thinkers.

If others are level-K thinkers, assume they believe everyone else is level-(K-1), who in turn model level-(K-2), etc., down to level-0, who always guess the midpoint of the allowed range. Use this chain of reasoning to predict the mean of all participants (*\textemdash*) including your own choice (*\textemdash*) and come up with your final response. Note that your own choice is part of the mean, so the optimal response solves for a fixed point.

Always write your final answer as: <response>YOUR_NUMBER_HERE</response>

(*\usertag*)

Including yourself, there are {N_PARTICIPANTS} participants in total.

All participants will be asked to pick a number between {MIN_RANGE} and {MAX_RANGE} inclusive. The winner will be the one choosing the number closest to 90% of the mean of all numbers chosen by all participants, including your own.

For example, in a game from 0 to 100, suppose you choose 14 and the other 10 participants choose 10, 22, 23, 30, 11, 18, 37, 18, 19, 20. The mean of all 11 numbers (including your 14) is (14+10+22+23+30+11+18+37+18+19+20)/11 = 222/11 (*$\approx$*) 20.18, and the target is 20.18 (*$\times$*) 90% (*$\approx$*) 18.16. The participants who chose 18 are the winners, as 18 is closest to 18.16.

In case of multiple winners, only one participant will be randomly chosen to win.

Now let's start this game. Please pick one number between {MIN_RANGE} and {MAX_RANGE} inclusive.
\end{promptbox}

\subsubsection*{Opponent-type framing (Table~\ref{tab:BCG-bias})}
\label{sec:appendix-prompt-bcg-bias}

Scenario \texttt{pBCG\_bias}.
$p = 2/3$, $n = 11$, range $= [0, 100]$.
Conditions: \texttt{OPPONENT\_TYPE} $\in$ \{human decision maker, game theory expert, large language model, artificial general intelligence agent, yourself, opponent\}.

\begin{promptbox}[pBCG Bias Prompt]
(*\systag*)

You are a participant in a one-shot economic decision task. You can assume that the other participants are {OPPONENT_TYPE}s.

Level-0 players are naive and always guess the midpoint of the range. Level-1 players believe others are level-0, level-2 believe others are level-1, and so on. Use this chain of reasoning to come up with your final response.

Your task is to choose the action that gives you the highest expected payoff based on what you believe the other participants will do by simulating a reasoning chain.

Always write your final answer as: <response>YOUR_NUMBER_HERE</response>

(*\usertag*)

There are {N_PARTICIPANTS} participants in total, including yourself. The other participants are {OPPONENT_TYPE}s.

All participants will be asked to pick a number between {MIN_RANGE} and {MAX_RANGE} inclusive. The winner will be the one choosing the number closest to 2/3 of the mean of all numbers chosen by all participants, including your own.

For example, if you choose 14 and the other 10 participants choose 10, 22, 23, 30, 11, 16, 37, 18, 19, 20, then you will be the winner: the mean of all 11 numbers (including your 14) is 220/11 = 20, the target is 20 (*$\times$*) 2/3 (*$\approx$*) 13.33, and 14 is closest to 13.33.

Or, if you choose 25 and the other 10 participants choose 90, 100, 50, 20, 60, 80, 5, 70, 10, 40, then the participant who chose 40 will be the winner: the mean of all 11 numbers is 550/11 = 50, the target is 50 (*$\times$*) 2/3 (*$\approx$*) 33.33, and 40 is closest to 33.33.

In case of multiple winners, only one participant will be randomly chosen to win.

Now let's start this game. Please pick one number between {MIN_RANGE} and {MAX_RANGE} inclusive.

Write your final answer as: <response>YOUR_NUMBER_HERE</response>
\end{promptbox}

\subsubsection*{Spontaneous reasoning traces (Table~\ref{tab:backtrack}, Figure~\ref{fig:bcg-claude})}
\label{sec:appendix-prompt-bcg-tracing}

Scenario \texttt{pBCG\_paper\_w\_tracing}.
$p = 0.9$, $n = 11$, range $= [1250, 8761]$.
No level-$k$ instruction; the structured output format elicits and exposes the model's spontaneous reasoning steps.

\begin{promptbox}[pBCG Tracing Prompt]
(*\systag*)

You are a participant in a one-shot economic decision task.

Your goal is to choose the number that maximizes your chance of winning.

Always respond strictly in the following structured format:

  <response>
  {
    "reasoning_steps": [
      {
        "step": 1,
        "reasoning": "Your reasoning clearly explained here.",
        "chosen_number": "NUMBER"
      }
      <!-- Add additional steps if you reconsider -->
    ],
    "final_decision": "NUMBER",
    "reflection": "Your optional reflection on your reasoning or decision"
  }
  </response>

(*\usertag*)

Including yourself, there are {N_PARTICIPANTS} participants in total.

All participants will be asked to pick a number between {MIN_RANGE} and {MAX_RANGE} inclusive. The winner will be the one choosing the number closest to 90% of the mean of all numbers chosen by all participants, including your own.

For example, in a game from 0 to 100, suppose you choose 14 and the other 10 participants choose 10, 22, 23, 30, 11, 18, 37, 18, 19, 20. The mean of all 11 numbers (including your 14) is (14+10+22+23+30+11+18+37+18+19+20)/11 = 222/11 (*$\approx$*) 20.18, and the target is 20.18 (*$\times$*) 90% (*$\approx$*) 18.16. The participants who chose 18 are the winners, as 18 is closest to 18.16.

In case of multiple winners, only one participant will be randomly chosen to win.

Now let's start this game. Please pick one number between {MIN_RANGE} and {MAX_RANGE} inclusive.

Important: Remember to use the required structured format output in <response></response>.
\end{promptbox}

\subsection{Money Request Game (MRG)}

\subsubsection*{Level-$k$ depth (Figure~\ref{fig:model-capability})}

Scenario \texttt{pMRG\_depth}.
Integer requests in $[11, 20]$ shekels; bonus of 20 shekels for requesting exactly one less than the opponent.
Conditions: \texttt{COMPETITION\_DEPTH} $\in \{\mathrm{L1}, \mathrm{L2}, \mathrm{L4}, \mathrm{L8}\}$.

\begin{promptbox}[MRG Depth Prompt]
(*\systag*)

You are a very smart decision-maker playing an economic game. You can assume that others participating in the game are level {COMPETITION_DEPTH} thinkers.

Level-0 players are naive and always request the maximum (20 shekels).
Level-1 players believe others are level-0, Level-2 believe others are level-1, and so on.
Use this chain of reasoning to come up with your final response.

Always write your final answer as: <response>YOUR_NUMBER_HERE</response>

(*\usertag*)

You and another player are playing a game in which each player requests an amount of money. The amount must be (an integer) between 11 and 20 shekels. Each player will receive the amount he requests. A player will receive an additional amount of 20 shekels if he asks for exactly one shekel less than the other player.

What amount of money would you request?

Remember, you are playing against a level {COMPETITION_DEPTH} thinker.

Write your final answer as: <response>YOUR_NUMBER_HERE</response>
\end{promptbox}

\subsubsection*{Opponent-type framing (Table~\ref{tab:BCG-bias})}

Scenario \texttt{pMRG\_bias}.
Same game; no level-$k$ instruction.
Conditions: \texttt{OPPONENT\_TYPE} $\in$ \{human decision maker, game theory expert\}.

\begin{promptbox}[MRG Bias Prompt]
(*\systag*)

You are a very smart decision-maker playing an economic game.
You can assume that others participating in the game are {OPPONENT_TYPE}s.

Always write your final answer as: <response>YOUR_NUMBER_HERE</response>

(*\usertag*)

You and another player are playing a game in which each player requests an amount of money.
The amount must be an integer between 11 and 20 shekels.
Each player will receive the amount he requests.
A player will receive an additional amount of 20 shekels if he asks for exactly one shekel less than the other player.

What amount of money would you request?

The other player is: {OPPONENT_TYPE}.

Write your final answer as: <response>YOUR_NUMBER_HERE</response>
\end{promptbox}

\subsubsection*{Emergence across model scale (Figure~\ref{fig:MRG-emergence-expert})}

Scenario \texttt{MRG\_emergence}.
Same game; expert condition only (\texttt{PLAYER} = \texttt{expert}).
Run across a wide range of model sizes to study the emergence of strategic behavior.

\begin{promptbox}[MRG Emergence Prompt]
(*\systag*)

You are a very smart decision-maker playing an economic game.
You can assume that others participating in the game are {PLAYER}s.

Always write your final answer as: <response>YOUR_NUMBER_HERE</response>

(*\usertag*)

You and another player are playing a game in which each player requests an amount of money.
The amount must be an integer between 11 and 20 shekels.
Each player will receive the amount he requests.
A player will receive an additional amount of 20 shekels if he asks for exactly one shekel less than the other player.

What amount of money would you request?

The other player is: {PLAYER}.

Write your final answer as: <response>YOUR_NUMBER_HERE</response>
\end{promptbox}

\subsection{Unlabeled Matrix Game (UMG)}

\subsubsection*{Level-$k$ depth (Figures~\ref{fig:model-capability} and~\ref{fig:UMG-expert})}

Scenario \texttt{ladder\_depth}.
Player~1 row player (actions A--F) vs.\ Player~2 column player (actions K--P).
Conditions: \texttt{COMPETITION\_DEPTH} $\in \{\mathrm{L1}, \mathrm{L2}, \mathrm{L4}, \mathrm{L8}\}$.

\begin{promptbox}[UMG Depth Prompt]
(*\systag*)

You are a participant in a one-shot strategic decision task. You can assume that the other player is a level {COMPETITION_DEPTH} thinker.

Level-0 players choose actions uniformly at random. You are Player 1 (row player), so the level-0 opponent (Player 2, column player) picks each of {K, L, M, N, O, P} with probability 1/6.
Level-1 players believe others are level-0, level-2 believe others are level-1, and so on. Use this chain of reasoning to come up with your final response.

Always write your final answer as: <response>YOUR_ACTION_HERE</response>

(*\usertag*)

You are Player 1 in a one-shot strategic game.

Your task is to choose the strategy that gives you the best expected payoff, anticipating how Player 2 will choose.

You are always Player 1 (row player), choosing from {A, B, C, D, E, F}.
Player 2 is always the column player, choosing from {K, L, M, N, O, P}.

For each pair (your action, their action), the first number is your payoff, the second is theirs.

Payoff table (Player 1):
        K   L   M   N   O   P
    A:  75  27  96  39   8  18
    B:  77  56  22  18  84  30
    C:  72  63  41  81  48  77
    D:  73  37  26  82  24  92
    E:  45  26  91  19  85  32
    F:  58  48  83  67  25  94

Payoff table (Player 2):
        A   B   C   D   E   F
    K:  75  77  72  73  45  58
    L:  27  56  63  37  26  48
    M:  96  22  41  26  91  83
    N:  39  18  81  82  19  67
    O:   8  84  48  24  85  25
    P:  18  30  77  92  32  94

E.g., if Player 2 chooses action M, and you (Player 1) choose D, the payoffs will be:
- Player 1: 26
- Player 2: 26

!!! Important !!!
Remember, you are playing against a level {COMPETITION_DEPTH} thinker.
Write your final answer as: <response>YOUR_ACTION_HERE</response>
\end{promptbox}

\subsubsection*{Opponent-type framing (Figure~\ref{fig:UMG-expert})}

Scenario \texttt{ladder-expert-eq}.
Same payoff matrix; no level-$k$ depth label.
Conditions: \texttt{PLAYER} $\in$ \{human, expert, opponent\}.

\begin{promptbox}[UMG Opponent-Type Prompt]
(*\systag*)

You are a very smart decision-maker playing an economic game. You can assume that others participating in the game are {PLAYER}s.

Level-0 players are naive and pick each of either {A, B, C, D, E, F} or {K, L, M, N, O, P} with probability 1/6.
Level-1 players believe others are level-0, Level-2 believe others are level-1, and so on.
Use this chain of reasoning to come up with your final response.

Always write your final answer as: <response>YOUR_ACTION_HERE</response>

(*\usertag*)

You are Player 1 in a one-shot strategic game.

Your task is to choose the strategy that gives you the best expected payoff, anticipating how Player 2 will choose.

You are always Player 1 (row player), choosing from {A, B, C, D, E, F}.
Player 2 is always the column player, choosing from {K, L, M, N, O, P}.

For each pair (your action, their action), the first number is your payoff, the second is theirs.

Payoff table (Player 1):
        K   L   M   N   O   P
    A:  75  27  96  39   8  18
    B:  77  56  22  18  84  30
    C:  72  63  41  81  48  77
    D:  73  37  26  82  24  92
    E:  45  26  91  19  85  32
    F:  58  48  83  67  25  94

Payoff table (Player 2):
        A   B   C   D   E   F
    K:  75  77  72  73  45  58
    L:  27  56  63  37  26  48
    M:  96  22  41  26  91  83
    N:  39  18  81  82  19  67
    O:   8  84  48  24  85  25
    P:  18  30  77  92  32  94

E.g., if Player 2 chooses action M, and you (Player 1) choose D, the payoffs will be:
- Player 1: 26
- Player 2: 26

!!! Important !!!
Remember, the other player is: {PLAYER}.
Write your final answer as: <response>YOUR_ACTION_HERE</response>
\end{promptbox}

\end{document}